\documentclass{article}

\usepackage[english]{babel}
\usepackage[letterpaper,top=2cm,bottom=2cm,left=3cm,right=3cm,marginparwidth=1.75cm]{geometry}
\usepackage{graphicx}
\usepackage{xcolor}
\usepackage{abstract}

\usepackage{here}
\usepackage{tabularx}
\usepackage[edges]{forest}
\usepackage{tikz}
\usepackage{tikz-qtree}
\usepackage{longtable}
\usepackage[whole]{bxcjkjatype}
\usepackage{hyperref}
\hypersetup{
    colorlinks=true,
    linkcolor=magenta,
    citecolor=magenta,
    urlcolor=magenta
}

\title{\bfseries Speculative Exploration on the Concept of Artificial Agents Conducting Autonomous Research}
\author{\normalsize \bfseries Shiro Takagi \\
\normalsize Independent Researcher\footnote{\href{https://t46.github.io/}{https://t46.github.io/} \\
The LaTeX file for this paper is currently hosted on GitHub. As the paper is still a work in progress, it may contain errors. Should you have any suggestions for improvement, or if you identify any mistakes or misunderstandings, please do not hesitate to submit a pull request to the repository. Your contributions towards refining this paper are greatly appreciated: \href{https://github.com/t46/research-automation-perspective-paper}{https://github.com/t46/research-automation-perspective-paper} }}

\date{}

\begin{document}
\sloppy
\maketitle

\begin{abstract}
This paper engages in a speculative exploration of the concept of an artificial agent capable of conducting research. Initially, it examines how the act of research can be conceptually characterized, aiming to provide a starting point for discussions about what it means to create such agents. The focus then shifts to the core components of research: question formulation, hypothesis generation, and hypothesis verification. This discussion includes a consideration of the potential and challenges associated with enabling machines to autonomously perform these tasks. Subsequently, this paper briefly considers the overlapping themes and interconnections that underlie them. Finally, the paper presents preliminary thoughts on prototyping as an initial step towards uncovering the challenges involved in developing these research-capable agents.
\end{abstract}

\tableofcontents

\section{Introduction}
Research has been the foundation of human progress. Through research, humans have deepened their understanding of the world and created unprecedented innovations, leading to groundbreaking advancements. It would be not an exaggeration to say that the future development of humanity heavily depends on the evolution and progress of research endeavors.

Since the inception of artificial intelligence (AI) research, a key goal has been to develop AI capable of conducting research \cite{zenil2023future}. AI-led research not only accelerates existing research and development but also offers significant potential for improving research practice and methodologies themselves, unencumbered by human cognitive limitations, research conventions, or unnecessary social constraints \cite{zenil2023future,kitano2021nobel}.

Researchers have developed systems that automate scientific decision makings \cite{lindsay1993dendral}, infer natural laws \cite{langley1987scientific}, autonomously cycle through hypothesis and experimentation \cite{king2004functional}, and many more \cite{zenil2023future,zenil2023}. With advancements in machine learning, there have been remarkable successes in using it for scientific discovery \cite{wang2023scientific,xu2021artificial,zhang2023artificial,ai4science2023impact}. These efforts by humanity have significantly advanced the potential of machines as valuable assistants in research.

On the other hand, the journey toward fully autonomous intelligent agents\footnote{
In this paper, the terms AI, machine, and agent are used interchangeably.
} capable of conducting research is still ongoing \cite{zenil2023future,coley2020autonomousII}. ``Autonomous'' here implies functioning without human intervention, prior design, or preparation. An agent is considered more autonomous if it can conduct research with less human involvement. By ``agents that conduct research,'' I refer to a single agent capable of performing research activities in various fields, like history, mathematics, or physics. The wider the range of research a single agent can handle, the more general it is considered. In this paper, when referring to an agent capable of conducting research, it denotes a general and autonomous artificial researcher. The realization of such agents has been a deeply held aspiration in the quest for advancing human research capabilities.

While there are excellent papers presenting perspectives on AI capable of conducting research \cite{zenil2023future,coley2020autonomous,coley2020autonomousII,kitano2021nobel,wang2023scientific,zenil2023,zhang2023artificial,hope2022computational,national2022automated}, there is still much to discuss about the nature of such agents and what it means to create an agent capable of doing research. Therefore, this paper presents a speculative thought around the concept of artificial intelligent agents capable of conducting research. This discussion aims to provide an opportunity to consider what future discussions are needed as we work towards realizing agents capable of conducting research 

First, I will explore conceptually how to characterize the research activities. This preliminary discussion is intended to help us consider what it would mean to create an agent capable of conducting research and what discussions would be necessary for its development. Next, I will discuss elements widely considered essential in research, specifically exploring the nature of question construction, hypothesis generation, and hypothesis verification, along with the potential challenges in autonomously performing these tasks.  Subsequently, I will consider topics that combine these elements, topics common to them, and points that could not be discussed previously. Finally, as a reference, I will share some simple, preliminary ideas for prototyping aimed at identifying challenges in developing an agent capable of conducting research.

The speculative discussion in this paper is still in its early stages and is provisional. Given the breadth of the subject matter and my limited capabilities, each point of discussion may be somewhat superficial or not entirely accurate. I plan to update these discussions continuously. Therefore, if anyone notices any points that should be improved, errors, or topics that would be beneficial to discuss, I would greatly appreciate your feedback.

\section{Conceptual Characterization of Research}
\label{section-what-is-research}

Understanding the fundamental nature of research is crucial for creating an agent capable of autonomous research. This section will therefore speculatively consider how the act of research can be characterized.

The aim of this section is not to establish a universal and singular definition of research, a task that exceeds the scope of this paper. Rather, it explores some characteristics of research to provide a provisional basis for discussions on the development of an artificial researcher. 

As such, the definition of research presented here should be regarded as tentative and operational, and the ensuing discussion is just one example of an endeavor to characterize research. Refining our understanding of research through in-depth discussions in the future is essential for the development of agents capable of conducting research. 

\subsection{Research as Knowledge Production}

While finding a unified, all-encompassing definition of research or science remains infeasible \cite{chalmers2013thing,sep-scientific-method}, various interpretations exist. For instance, one view posits that research occurs ``whenever we gather information to answer a question that solves a problem'' \cite{booth2003craft}, while another describes research (and development) as comprising ``creative and systematic work undertaken in order to increase the stock of knowledge'' \cite{manual2015guidelines}. Additionally, some perceive the science as ``processes that maximize the evidence for a generative model of the sensed and measured world'' \cite{balzandistributed}. These descriptions highlight different crucial aspects of research, with none being entirely incorrect or absolutely definitive.

Among these, a broadly recognized definition can be that \textbf{research is an endeavor to generate new knowledge}. Since this characterization seems to align with our research practices, regardless of the field, this definition serves as a suitable starting point for our discussion. In this paper, I adopt this interpretation as a provisional working definition. Specifically, I will regard research as the attempts to produce new knowledge for certain society. The inclusion of ``for certain society'' acknowledges the societal relativity of knowledge, a point I will elaborate on in subsequent sections.

\subsection{Knowledge Production as Belief Revision}
\label{section-knowledge-production-as-belief-revision}
Having defined research as the endeavor to generate new knowledge, it becomes important to consider what ``knowledge'' itself entails, and what constitutes its production. This section aims to explore these concepts.

Defining knowledge and the process of knowledge production rigorously remains an unsettled philosophical debate \cite{sep-epistemology}. Given that providing a precise definition of knowledge is beyond the scope of this paper, an in-depth exploration of these debates will not be undertaken here. Instead, this section aims to present a basic and preliminary conception of what knowledge might entail so that it serves as a starting point  for further discussion.

\subsubsection{Knowledge as Belief}
The concept of knowledge has long been a subject of debate within \textit{epistemology}, a branch of philosophy. This paper will reference some basic and introductory concepts in this field as an exemplar to explore how research might be characterized.

Within epistemology, knowledge has traditionally been viewed as \textit{justified true belief (JTB)} \cite{sep-epistemology}. The term ``true'' is challenging to define rigorously; however, for the purposes of this discussion, it can be understood as something that corresponds with fact. ``Belief'' is tentatively defined as an individual's thought or conviction about a subject. ``Justified'' implies that it is reasonable to hold such a belief. The nature of justification has been a focal point of debate in epistemology, particularly following criticisms that JTB may not adequately define knowledge \cite{gettier1963justified}. Consequently, the refinement or expansion of the JTB has become a significant topic in epistemological discourse \cite{sep-epistemology}.

Although many philosophers contend that the JTB properties alone are insufficient for defining knowledge, there is some consensus that they might be necessary components \cite{sep-epistemology}. In epistemological debates, rather than discarding JTB entirely, many theorists use it as a foundational concept. Hence, this paper will tentatively adopt JTB as a preliminary basis for discussion. 

The subsequent sections will explore how research is conceptualized under this definition. Analyzing the congruences and discrepancies between the implications drawn from this characterization and our expectations of a research-capable agent will generate insights for refining the definition of research and hypothesizing the capabilities such agents should possess.

\subsubsection{Knowledge Production as Belief Updating}
In the current framework, knowledge is equated with belief. Therefore, knowledge production can be reinterpreted as the process of adopting the belief that a particular proposition is true and subsequently revising this belief based on justification. Essentially, in this framework, knowledge production equates to the updating of beliefs.

While it might initially seem counterintuitive to view research as a process of updating beliefs, this perspective gains plausibility when considering several factors. Research involves the continual revision of hypotheses and theories; inductive reasoning, unlike deduction, does not conclusively prove propositions; and, as will be discussed, knowledge is essentially subjective. The characterization of research as belief aligns well with these aspects of research. Therefore, this characterization seems reasonably valid.

The primary aim of research can be conceptualized as uncovering the unknown truths of the world. Consequently, the justification employed in research must be capable of accurately discerning the truth or falsity of propositions. This form of justification, known for its capacity to lead to truth, is termed ``truth-conducive.'' While there are varied debates on the nature of justification, it is widely accepted that justification in research should indeed be truth-conducive. Thus, knowledge production can be conceptualized as the construction of belief in new propositions about the world and ascertaining their veracity through truth-conducive justification.

\subsection{To Know Depends on Knowing Subjects}

Fundamentally, the concept of knowing presupposes not only the existence of the object being known but also of the subject doing the knowing. This interplay explains why the definition of knowledge incorporates the subjective element of belief. Consequently, while the notion that knowledge is a form of belief might initially seem counterintuitive, it holds validity in this context. 

Moreover, conceptualizing research as an updating of beliefs aligns closely with actual research practices. For instance, the experimental validation of a hypothesis reinforces our belief in its truth or falsity. Our confidence in a hypothesis increases as it withstands various rounds of verification. This process of iterative validation and belief reinforcement mirrors the concept of research as a continual renewal of beliefs.

Finally, since the justification in research is expected to be truth-conducive, the knowledge thus produced would have an objective quality. Therefore, in conjunction with the discussion in the previous section, it seems that the use of the subjective concept of belief is not that problematic.

\subsubsection{Knowledge for Humans}
As previously discussed, research is a pursuit dedicated to uncovering the unknown truths of the world, necessitating that the knowledge it generates be novel. This raises the question: what constitutes new or unknown knowledge?

Under the current definition, knowledge is a justified belief regarding the truth or falsity of a certain hypothesis. Thus, unknown knowledge could be a state where such a belief is either non-existent or, if it exists, lacks justification. In simpler terms, within this framework, the state of certain knowledge being unknown is essentially a state of belief.

Given that the concept of knowing is contingent on the knowing subject, the notion of the unknown is also inherently subject-dependent. In the realm of research, the term ``subject'' has seemingly encompassed humanity at large. Researchers do not deem knowledge as unknown simply because it eludes an individual; it is regarded as truly unknown only when it is beyond the collective understanding of humanity. Consequently, the knowledge produced through research is expected to contribute to the collective understanding of human society. This is the reason why the term ``for society'' is included in the definition of research.

\subsubsection{Knowledge for Non-Humans}

Since the act of knowing is subject-dependent, it is theoretically feasible to conceive of non-human knowledge and research by considering non-human agents as knowing subjects. Naturally, there is also the unknown for these non-human agents. Such knowledge and unknowns for non-humans can naturally differ from those for humans. Therefore, when referring to ``knowledge,'' ``unknown,'' or ``novel,'' it is necessary to specify for whom these concepts apply.

While this discussion might seem like mere speculation, the idea that machines might have a different scope of the unknown compared to humans has implications for realizing an artificial researcher. This is because it suggests that merely replicating current research methodologies in AI might not necessarily yield new knowledge for humans.

Research methods essentially have been developed to uncover truths unknown to knowing subjects. Therefore, an AI mimicking these methods might only reveal truths unknown to itself, which may not align with human knowledge gaps. If we want AI to conduct research autonomously, we must find a way to ensure it understands what is unknown to humans, not just to itself, and guide it to discover knowledge that is truly unknown in the human context.

This is just a preliminary discussion. It is hoped that further discussion will continue on how to realize them for developing research-capable AI

\subsubsection{Can Truth-Conducive Justification Be Developed by Non-Human Agents?}

I acknowledge that we cannot definitively ``prove'' our empirical verification methods to be entirely truth-conducive. Yet, given the myriad discoveries achieved through these methods, questioning their legitimacy seems to have little merit. If agents can thoroughly grasp and effectively apply these human-developed justifications, it is poised to uncover numerous unknown truths, a prospect few researchers would dispute.

Then, what about when the agent constructs its own methods of justification? Is it possible for an agent to construct new truth-conducive justification methods on its own, instead of just mastering human-developed methods? Humans have devised tools like statistical hypothesis testing to evaluate hypotheses; can AI similarly innovate unique methodologies? Even if it were possible, how significant would those be?

Our justification methods are founded on various premises, implying diverse interpretations of ``what constitutes a truth-conducive justification'' or ``what justification entails.'' These interpretations lead to multiple methods of justification even among humans \cite{otsuka2022thinking}. Consequently, when allowing artificial agents to autonomously develop justification methods, these agents must consider what can serve as justification, navigate value judgments regarding the nature and effectiveness of justification, and thereby conceive and select optimal methodologies.

This inquiry goes beyond mere philosophical speculation. Since the current justification methodologies have not been proved to be the optimal, the potential for superior methodologies exists in theory. Machines, unfettered by human cognitive limitations, theoretically possess the possibility to discover such methods. Importantly, truth-conducive verification methods does not essentially require human value judgments to evaluate their quality, thus, there is ample potential for machines to ``autonomously'' devise them. The feasibility, realization, and significance of these possibilities remain open for exploration and debate.

\subsection{Conclusion}
In this section, I have discussed a provisional working definition of research. My initial premise was the intuitive belief that research is an endeavor aimed at generating new knowledge for a particular society. The discussion subsequently delved into a speculative inquiry into the idea that that knowledge is fundamentally a form of belief and that the production of knowledge is the updating of beliefs. Building on these ideas, I presented some conjectural insights of non-human agents doing knowledge production.

It is important to note that the definition proposed here serves merely as a starting point. A more comprehensive and nuanced understanding of research can be cultivated through the collective insights of philosophers, scientists, and practitioners across various fields. This collaborative approach will enable us to delve deeper into the definition of research and develop more robust and effective guidelines to realize an artificial researcher.

\section{Question, Hypothesis, and  Verification}
\label{section-question-hypothesis-verification}
In the preceding section, I briefly examined a preliminary conceptual definition of research and its implications. This section shifts focus to the widely acknowledged fundamental components of research: question construction, hypothesis generation, and hypothesis verification. 

The objective here is to advance the discussion beyond the somewhat too abstract considerations in the previous section. By dissecting these core elements, this section seeks to offer a more concrete exploration of what constitutes research and the prospects of machines engaging in research activities. 

\subsection{Question Construction}
\label{section-question-construction}
The first essential element in research is \textit{question construction}. To produce new knowledge, it is imperative to recognize what is unknown and strive to generate that elusive knowledge. This act of identifying the unknown for investigation can be considered as the process of questioning. Subsequently, formulating potential answers for these questions constitutes hypothesis generation. Essentially, research can be reinterpreted as the act of posing and answering to questions. Furthermore, since research inherently involves lots of uncertainty, the generation of multiple questions, beyond the initial research question, is a natural part of the process. That is, in the pursuit of confronting the unknown, question construction is an inevitable aspect of research.

There have been the studies to find research questions and challenges from academic literature \cite{lahav2022search,oppenlaender2023mapping,surita2020can}, generate ideas for future work \cite{wang2019paperrobot}, and identify research trends \cite{krenn2020predicting,krenn2022predicting}. However, enabling machines to autonomously generate research questions is less common. In the domain of question answering from natural language processing (NLP) research, tasks exist for generating questions \cite{pan2019recent,zhang2021review}, but these are motivated differently from generating research questions. Research on artificial curiosity for generating non-textual questions has been conducted \cite{schmidhuber1991possibility}, yet it doesn't generate research questions akin to a human researcher, as far as I know. Recent advancements in large language models (LLMs) have led to initial attempts at generating research questions \cite{liu2023creative,lahat2023evaluating}, but this area remains nascent.

While there are efforts towards automation as shown above, the number of these attempts are relatively limited compared to those for hypothesis generation and verification. The automation of question construction, or determining the underlying goals of such automation, is recognized as a key challenge in the field of research automation \cite{coley2020autonomousII,zenil2023,kitano2021nobel}. 

In this section, I start with a speculative exploration of the nature of questioning. This will be followed by a discussion of the open challenges in enabling an artificial agent to effectively pose research questions.

\subsubsection{What is Questioning?}
Asking questions is often characterized as an information-seeking behavior \cite{watson_2021,taylor1962process}. This behavior typically involves two distinct steps: firstly, recognizing an \textit{information need}, and secondly, undertaking actions to acquire the desired information \cite{wilson1997information,case2016looking}. While not all information-seeking behaviors necessitate linguistic expressions \cite{watson_2021}, in the context of research, queries are typically formulated in text. This textual formulation occurs between the stages of information need recognition and the initiation of information-seeking behavior. Specifically, in research, the process of question construction is generally understood as the journey leading to the formulation of such queries. Thus, for the purposes of this paper, question construction is regarded as the process culminating in the formulation of a query. The subsequent steps of information seeking are considered part of hypothesis generation and verification.

Recognizing an information need seems to involve at least two sub-processes: identifying the knowledge gap and deciding to address it (judging that the missing information is a ``need'').\footnote{
In this discussion, the process of recognizing an information need is described as initially identifying something as unknown and then deciding whether to formulate a question about it. However, the sequence of these steps is not fixed. For instance, one might first have a desire to know something and only afterward ascertain that it is indeed unknown. The critical aspect is that the process encompasses these two elements.
} Therefore, to enable an artificial agent to autonomously construct questions, it is necessary to consider how to imbue it with these capabilities. The following sections will delve into a speculative consideration and exploration of these steps.\footnote{
It is important to note that questions in research are not personal but societal in nature. This societal aspect may introduce slight variations in the question construction process. This point will be revisited later in the paper.
}

\subsubsection{Recognizing Unknowns}
Recognizing that certain knowledge is unknown typically involves an initial attempt to access that knowledge. This process usually entails referring to our personal knowledge base and, upon not finding the information, deeming it as unknown. For an individual, this knowledge base is essentially the memory stored within the brain. However, in the realm of research, the unknowns that researchers aim to elucidate are those unknown to a specific society, not just to an individual researcher. That is, a research-capable agent does not need to judge whether it is unknown to itself, but rather it can directly determine whether it is unknown to certain society. In this context, the knowledge base extends beyond the agent's memory to encompass societal knowledge sources, such as a collection of research papers.\footnote{
As mentioned earlier, something being unknown implies either the absence of a proposition or the presence of an unverified belief. Therefore, accurately determining the unknown from academic papers requires an assessment of whether each paper has been appropriately justified, meaning whether its verification processes are sound.
}

As outlined in Section \ref{section-what-is-research}, for machines to generate new knowledge beneficial to humans, they must be capable of identifying what is unknown to humans, not to themselves. While this task might initially seem as straightforward as conducting a literature survey as humans do, the reality may necessitate more complex approaches. The specific requirements for achieving this goal merit further discussion.

An additional consideration arises regarding the reliance on a machine's judgment to determine what is unknown to us. If an AI has been pre-trained on an extensive corpus of scholarly papers, its judgment might appear credible. However, as previously mentioned, an AI's determination of unknowns does not necessarily coincide with human unknowns. Therefore, it is not certain whether it is appropriate to unconditionally believe that what AI has determined to be unknown is indeed unknown. This issue becomes increasingly pertinent and significant as AI amass more knowledge and enhance their capabilities.

\subsubsection{Deciding What Knowledge to Seek}
\label{section-deciding-what-knowledge-to-seek}
While we encounter numerous unknowns, we do not formulate questions for each one, as not all unknowns hold equal ``importance'' or ``interest.'' Instead, we construct questions for matters we are eager to understand. This process involves assessing the ``value'' of questions based on certain criteria to determine their worthiness of pursuit. For individuals, this can be a largely subconscious process. However, in research, this need not be an internal process, as long as that is the value judgments of questions.

Knowledge, in itself, is value-neutral. The ``value,'' ``significance,'' or ``goodness'' of knowledge is ascribed by its users. A noteworthy aspect here is that the criteria for determining ``value'' are subjective and arbitrary. Hence, if we aim for artificial agents to autonomously pose questions meaningful to humanity, it is crucial to identify what constitutes ``good'' or ``significant'' questions for us and instill these values in the agents.

On the other hand, it is also vital to recognize that certain questions deemed ``unimportant'' by us may actually hold importance under different criteria. Fundamental research, for example, often yields knowledge initially perceived as ``useless'' but later proves pivotal for innovations. Human cognitive limitations may sometimes hinder our ability to fully appreciate the potential utility of such knowledge. Moreover, social factors unrelated to the initial purpose of knowledge production can influence our value judgments, implying that these human judgments are not always optimal.

Given that machines are not inherently limited by such constraints, they could theoretically make more effective value judgments. Therefore, while providing some guidance to ensure that the generated question is relevant to humans is crucial, developing agents capable of autonomously constructing these value criteria themselves may also be fruitful. How to achieve this forms a significant open challenge in developing research-capable agents.\footnote{
Kitano has described the approach where humans apply their value judgment criteria to determine questions and hypotheses as \textit{value-driven science} \cite{kitano2021nobel}. He advocates for the advancement of \textit{exploration-driven science}, which prioritizes extensive and comprehensive exploration. While a completely value-neutral exploration is unattainable, the notion of employing diverse and extensive criteria is indeed significant for the future of research. By embracing a wider range of criteria, we can expand the exploration space of knowledge.
}

\subsubsection{Origin of Information Need}
I previously outlined that question formation begins with the recognition of an information need. This leads to the question: what triggers the recognition of an information need?

The initiation of this process can be attributed to various factors. Some researchers may generate questions through logical contemplation aimed at achieving a specific goal. Others might identify questions upon noticing anomalies in experimental data or inconsistencies between theoretical assumptions and actual observations. Researcher also sometimes search questions that can be answered by techniques that you have. Furthermore, humans typically do not rely on a singular criterion for value judgment. Instead, multiple criteria are often intricately combined and weighted according to the context, culminating in a complex value assessment process. To develop an agent capable of autonomously constructing research questions as humans do, therefore, it appears necessary to create a system with a general methodology for questioning applicable across these diverse scenarios.

The process in humans that connects these various triggers to an information need, and the development of an agent capable of emulating this process, remains an open question. Researchers in the field of curiosity, which is broadly conceptualized as a ``drive state for information'' \cite{kidd2015psychology}, have been investigating this challenge. Curiosity is often characterized as a precursor to information need in information-seeking processes \cite{case2016looking}. In reinforcement learning, efforts to instill curiosity or knowledge-based intrinsic motivation in AI have been explored. Here, curiosity is defined in terms of novelty, information gain, or prediction error, and is considered a catalyst for exploration \cite{aubret2019survey}.

These efforts provide insights into implementing mechanisms that drive AI towards question formation. However, we are still distant from realizing a system that autonomously constructs research questions under complex value judgments, as humans do. A significant challenge lies in identifying the minimal input required for question generation; namely, while the minimal input is clear for hypothesis generation and hypothesis verification, it remains unclear for the construction of questions. Designing a complex, contextually adaptive internal driving force for questioning remains a significant hurdle. Identifying the prerequisites for an AI system with such a mechanism is an ongoing challenge.

\subsubsection{Examples of Criteria for Evaluating Research Questions}

Thus far, I have discussed abstract concepts related to questioning in general. Now, I will move on to focusing specifically on the characteristics pertinent to research questions.

The ``quality'' of a research question can be evaluated against various criteria. Here, I will briefly explore some examples to illustrate how humans seemingly appraise the value of a research question. It is important to note that these examples are only a few of the many criteria utilized and do not represent a comprehensive list. 

One widely accepted criterion within the research community is that a question is important if it offers new perspectives, understandings, or conceptual advances, particularly those that challenge our common assumptions. For example, Alvesson and Sandberg emphasize the significance of such questions and discuss strategies for their construction \cite{alvesson2013constructing}. This criterion rests on the idea that a valuable question is one that significantly impacts our current knowledge. This seems to be a value that aligns with the highest-level objectives of the endeavor of research.

No matter how significant a question may be, if it is nearly impossible to address with current technology, deriving meaningful research outcomes from it may be unfeasible. Consequently, the feasibility of answering a question is considered a vital aspect of its quality \cite{hulley2007designing,alon2009choose,huntington2021effect}. Assessing feasibility involves complex decision-making, taking into account factors like available resources, researcher capabilities, deadlines, and technological constraints. An agent engaged in research would need the capacity for such multifaceted evaluations.

Another prevalent view is that research questions should stem from individual intellectual curiosity. Given that curiosity drives exploration \cite{oudeyer2018computational}, curiosity-driven research can foster exploration in the knowledge space. Research can be seen as an exploration of the world's truths, making this value standard important. However, curiosity is not the sole criterion for exploration; there may be better criteria for uncovering unknown truths. If agents can adopt such criteria, it might surpass human efficiency in uncovering truths.

In contrast to bottom-up curiosity-driven research, questions that contribute to achieving specific top-down goals are also considered valuable. For example, in corporate or government-led research, questions aligned with predetermined objectives are prioritized. Since we expect that agents capable of autonomous research will contribute to human-set goals, ability to make such value judgments deemed important.

In practice, the value of a research question is determined by integrating multiple criteria. Hulley et al. suggest that questions which are feasible, interesting, novel, ethical, and relevant (FINER) are considered valuable \cite{hulley2007designing}. Huntington-Klein argues that a good research question is one that is answerable and whose answer enhances our understanding of the world \cite{huntington2021effect}. As mentioned above, autonomous agents are also expected to determine the questions they should pursue based on such complex value judgments.

As emphasized, these criteria represent only a portion of the value judgments humans make in question formulation. Future discussions should further investigate the nature of these judgments, their role in scientific discovery, and how they can be replicated in artificial researcher.

\subsection{Hypothesis Generation}

The second integral element of research is hypothesis generation. Research inherently involves posing questions and endeavoring to answer them. Typically, researchers bifurcate the answering process into two phases: generating hypotheses and verifying them. Hypothesis generation entails predicting the answer to a posed question, while hypothesis verification involves examining the plausibility of that prediction. This two-stage approach is adopted because researchers address questions to which no one in this world knows the answer, making it challenging to immediately ascertain definitive answers. Therefore, the separation of hypothesis generation and verification represents a human-developed methodology for uncovering truths in a context of high uncertainty.

Hypothesis generation is often seen as a showcase of human creativity in research. The long-standing belief that human creativity defies analysis has led to the assumption that both question construction and hypothesis generation are inherently unanalyzable \cite{sep-scientific-discovery}. However, efforts to characterize this creative process began to emerge in the mid-20th century. Notable concepts include the role of abduction in generating hypotheses for ``why'' questions \cite{hanson1965patterns,magnani2011abduction}, the significance of analogical reasoning \cite{gentner2002analogy}, the interpretation of scientific discovery as a form of search problem \cite{langley1987scientific}, and the conceptualization of hypothesis generation as probabilistic sampling \cite{dasgupta2017hypotheses}.

The potential for machines to generate hypotheses has been a focal point in artificial intelligence research. Pioneering attempts to develop machines capable of hypothesis generation date back to the early 20th century \cite{langley1987scientific,lindsay1993dendral}. By the mid-2000s, advancements led to the creation of machines capable of making autonomous scientific discoveries \cite{king2004functional}.

\subsubsection{Hypothesis Generation and Machine Learning}
Generating hypotheses is predicting answers to questions from existing knowledge. This process essentially aligns closely with machine learning, particularly question-answering. As it involves predicting answers that even nobody knows, it can also be viewed as a prediction under significant distribution shifts.

Indeed, machine learning have become increasingly prominent in scientific hypothesis generation \cite{xu2021artificial,zhang2023artificial,wang2023scientific}. Attempts such as predicting protein structures \cite{jumper2021highly} and new materials \cite{merchant2023scaling} are all examples of hypothesis generation.\footnote{
Since there are a vast number of studies, I will skip the introduction of individual studies here.
}

The sources for hypotheses, the nature of the hypothesis space, and the representation of hypotheses differ across research fields. For instance, hypotheses can be represented combinatorially, and machines can be employed to explore these spaces to find hypotheses \cite{coley2020autonomous}. Some studies represent hypotheses as symbolic equations and try to discover them from scientific data \cite{kramer2023automated}, while others endeavor to generate or extract textual hypotheses from academic papers \cite{kang2022augmenting,chan2018solvent,wang2023learning,xu2023exploring,yang2023large}. The advent of LLMs has spurred efforts to generate hypotheses from the models' internal knowledge, without relying on direct information from academic papers \cite{park2023can,ai4science2023impact}.

While the specifics of hypothesis generation vary, a unified description can be drawn from certain perspectives. Viewing the hypothesis space as a human-defined and fixed entity, scientific discoveries can often be framed as search problems \cite{coley2020autonomous}. 
Since the hypothesis space is often combinatorially vast, strategies for efficient exploration in the space deemed necessary \cite{coley2020autonomousII,zenil2023future}, and efforts have been made to optimize exploration with techniques such as active learning. As another perspective, Wang et al. provide a categorization of how AI is utilized in scientific hypothesis generation, highlighting its applications in black box prediction, aiding hypothesis space exploration, and finding solutions within a differentiable hypothesis space \cite{wang2023scientific}.

As such, integration of machine learning to hypothesis generation has progressed significantly compared to question formulation and hypothesis testing. The applications in this field are vast and diverse, and detailed evaluation of individual cases goes beyond the author's capacity. Therefore, this paper does not provide introductions to specific cases. Those interested are encouraged to refer to survey papers in their respective fields.

While machine learning's application in hypothesis generation is notable, the development of AI capable of generating complex hypotheses in response to varying questions remains a challenge. Achieving such capability may require abilities to generate hypothesis in versatile and flexible manner and to construct hypothesis spaces themselves. Future discussions are expected to further explore how to realize these capabilities in AI.

\subsubsection{Speculation on Key Aspects of Hypothesis Generation by Machines}
It can be said that autonomous hypothesis generation in general manner by AI has already gained considerable attention, compared to question generation and hypothesis verification. That's largely because, as previously mentioned, predicting answers to questions is a problem already central to many machine learning researchers. Therefore, many challenges in aiming for AI that generates hypotheses as flexibly as humans overlap with the challenges of pursuing an artificial general intelligence (AGI). These include systematic thinking such as deduction, out-of-distribution generalization, causal inference, efficient exploration, and problem decomposition, all crucial for autonomous flexible hypothesis generation and considered fundamental in the pursuit of AGI as well.

In this section, I will preliminary explore elements deemed important for AI's ability to generate hypotheses. However, due to the circumstances mentioned in the previous paragraph, this discussion might intersect with existing debates in the realm of AGI, potentially lacking novelty. Nonetheless, I will explore two aspects that appear vital for the development of an artificial hypothesis generator.

Firstly, it's important to note that even AI might not know the answers to the research questions. It's not always true that a question unknown to humans is also unknown to machines, as has been repeatedly emphasized. However, once the answer is unknown to both humans and the machine, hypothesis generation can be a challenging task even for AGI. I acknowledge that this essentially  reduces to a problem of out-of-distribution generalization, but it's particularly challenging because no agents in this world know the answer. To solve such problems, machines, like humans, may need to recognize their ignorance of the answer, reduce uncertainty step-by-step, and gradually approach the answer. Current AI still does not even understand what it doesn't know \cite{guo2017calibration,maynez2020faithfulness}. How AI capable of reasoning under such high uncertainty can be realized remains an open question.

Secondly, the role of mathematics in hypothesis generation cannot be overstated. The first point to note is that the power of mathematics in hypothesis generation lies significantly in its deductive nature. Deduction ensures that if the premises are true, the resulting conclusions are also true, even if they may seem counterintuitive. This aspect gives AI, which largely depends on experiential inferences, a substantial advantage. Furthermore, as humans do by the hypothetico-deductive method, deduction enables the evaluation of hypotheses that are not directly testable. If deductive results are rejected, the hypothesis is deemed false; acceptance, conversely, strengthens its plausibility. This plays a crucial role in expanding the empirical knowledge boundaries in research. 

The abstract nature of mathematics is also important. Since ancient times, even before the formalization of deductive methods, mathematics engaged with concepts such as numbers, which are fundamentally abstract and have long captivated human interest \cite{david2010history}. The introduction of symbolic representation and manipulation has further amplified its abstract nature. Significantly, mathematics not only abstracts real-world objects but also engages in a cycle of further abstraction. By abstracting already abstracted concepts, it has developed highly sophisticated systems \cite{bochner1968role}. This level of abstraction allows for the reference to subjects not directly experienced, facilitating the progress in science \cite{heisenberg2008abstraction}. 

These characteristics render mathematics an indispensable tool in the process of hypothesis generation. AI capable of doing mathematics has not yet been realized, but related research attempts have made steady progress \cite{rabe2021towards,imani2023mathprompter}.

While these two elements are discussed separately, systematic thinking seems necessary for both, reinforcing the widely acknowledged importance of systematic or high-level thought in AI development. However, due to the extensive existing discourse on this topic \cite{goyal2022inductive}, I will not delve deeper into it here.

Due to my limitations, this paper only scratches the surface of this topic. I would appreciate any feedback from those with insights into elements not widely recognized in the machine learning community but deemed essential for autonomous hypothesis generation.

\subsection{Hypothesis Verification}
\label{section-hypothesis-verification}
The final critical element in the research process is the verification of hypotheses. We justify our belief in the truth or falsehood of a hypothesis by confirming the plausibility of our prediction in response to a question through verification. Thus, verification is essential for generating knowledge.

Verification hinges on the nature of the question and hypothesis posed. For instance, a ``why'' question demands verification methods that elucidate causal relationships. Questions about the physical world require interaction with physical world for verification. In cases where hypotheses are amenable to mathematical proof, such proof constitutes verification. This necessitates an agent capable of verification to possess an understanding of what constitutes verification and to develop suitable verification methods tailored to the specific question and hypothesis.

While there has been extensive discussion on AI in hypothesis generation, its involvement in verification is less explored. Certainly, some studies have utilized AI in aspects of verification, such as experimental design \cite{chaloner1995bayesian} and scientific simulations \cite{baker2019basic}. However, initiatives enabling AI to fully comprehend and independently execute verification processes akin to human scientific research are still limited. 

In machine learning, research focusing on the validation of scientific claims \cite{wadden2020fact}, factual accuracy of predictions \cite{guo2022survey}, evidence search to support hypotheses \cite{koneru2023can}, and self-verification of machine responses \cite{dhuliawala2023chain} aligns with aspects of verification. Nevertheless, none of them aim to construct and execute verification as humans do in a scientific research. The automation of peer review \cite{kousha2022artificial,lin2021automated1} is also related to verification in the sense that it demands judgment on the validity of the verification, but it does not generate verification.

In this section, I aim to delve into the concept of verification to stimulate further contemplation. Having already addressed the nature of verification, or justification, in Section \ref{section-what-is-research}, I will omit that discussion here. Instead, I will focus on experimentation, an essential aspect of human-like verification in scientific inquiry.

\subsubsection{Experimentation}
\label{section-experimentation}
No researcher would deny the importance of experiments. An experiment involves the planning and execution of a series of procedures to empirically test a hypothesis, essentially constituting the process of verification in empirical science. Therefore, any agent capable of verification must necessarily possess the ability to conduct experiments.

In experiments, phenomena that are difficult to observe, or the effects of various conditions, are precisely investigated. This is achieved by artificially generating phenomena in a controlled manner and actively intervening in them \cite{radder2009philosophy}. Such interventions create the different observations of interest. These observations are recorded as experimental data, and subsequent analysis of the data determine the validity of the hypothesis.\footnote{
Experiments are not conducted solely during the verification phase but also when generating hypotheses. Furthermore, new questions and hypotheses are often formulated based on the results obtained from experiments. In these instances, the process leading up to data generation, or conducting experiments not solely for verification but for data generation and some form of data analysis, seems to be what is referred to as an experiment. This paper defines an experiment as the planning, preparation, data generation, analysis, and determination of verification results. However, be aware that this definition may not always reflect actual practice.
}

To conduct an experiment, one must first design it, document the procedures, and plan its execution. Planning requires an understanding of what constitutes a successful hypothesis test and the ability to devise methods to realize this using existing technology. For example, if the hypothesis pertains to the causes of a particular phenomenon, one must understand what causality is in the first place and how it can be identified in order to plan the verification process. 

Preparations for the experiment are also essential. These preparations can include purchasing chemicals, preparing flasks, training animals, applying to ethics committees, constructing necessary equipment, and sometimes even building large apparatus like accelerators from scratch. Unfortunately, since research aims to uncover the unknown, constructing equipment from scratch for experiments is not uncommon in research. The autonomous execution of these preparations by a non-human agent from scratch seems almost infeasible.

After the preparation for the experiment is complete, the experiment is conducted according to the experimental protocol. This task also presents considerable challenges for autonomous machines. The reason is that even a single experiment requires a myriad of low-level operations such as grasping, cutting, carrying, mixing, moving, pouring, dispensing, washing, and opening lids. These operations need to be flexibly combined and executed according to the self-generated experimental protocol. An autonomous machine capable of conducting experiments must possess the ability to generate these operations flexibly in response to the experimental protocol.

\subsubsection{Automating Experimentation}
As we observe, the challenge of making a machine fully autonomous in planning, preparing, and executing experiments is considerable. Particularly, since the specific experiments to be conducted cannot be determined until questions and hypotheses are formulated, enabling a machine to autonomously conduct research from question construction demands the capability to accommodate many possible experimental scenarios. This, I believe, represents one of the greatest barriers to creating machines capable of autonomously conducting research.

Automating experiments is a daunting task, yet humanity has made steady progress in this area. In relation to the planning stage of experiments, the automation of exploring experimental conditions have a long-standing history, for example. Wang et al. have summarized these studies, which utilize AI to assist in experiment planning, research guidance, and generating observational data through numerical simulations \cite{wang2023scientific}.

Furthermore, there is an initiative known as \textit{laboratory automation} or \textit{self-driving lab} that aims to automate experiments, including their execution – an aspect previously mentioned as challenging \cite{holland2020automation,abolhasani2023rise}. A notable example is the research in genetics by King et al., who fully automated the cycle of hypothesis generation, verification, and the discovery of new hypotheses \cite{king2004functional}. Another example is the work of A.I. Cooper, which facilitated the use of experimental equipment by autonomous robots, similar to human researchers \cite{burger2020mobile}. These are just a few examples, and there is a vast number of studies in this field.

These examples illustrate efforts to autonomously drive the research cycle, encompassing hypothesis generation, planning and execution of experiments, and generation of new hypotheses based on experimental results. Such endeavors are referred to as the \textit{closed-loop} automation of scientific discovery \cite{zenil2023future}, representing a significant milestone in achieving high autonomy in research automation. Additionally, there are efforts to develop humanoid robots capable of conducting multiple different experiments with a single robot, considered a foundational step towards more generalized research automation \cite{yachie2017robotic}.

In recent years, there have been efforts to explore possibilities of autonomous experiment using LLMs \cite{boiko2023emergent,qin2023gpt,charness2023generation}. For instance, Boiko et al. developed an autonomous agent comprising multiple LLMs that successfully designed and executed complex scientific experiments \cite{boiko2023emergent}.

While we have primarily discussed experimental data generation process,  validation also requires interpretation of the data. Observation inherently involves theoretical underpinnings \cite{hanson1965patterns}. Hence, interpreting experimental data necessitates adequate prior knowledge. Some studies are focused on enabling machine learning models to interpret scientific data by embedding physical prior knowledge, like symmetries, differential equations, and intuitive physics, into them \cite{hao2022physics,karniadakis2021physics}.\footnote{
Interpretation of scientific data is not solely for validation purposes. Thus, these technologies extend beyond just automating validation processes.
}

Various research efforts have significantly advanced the automation of experiments. However, it is also true that numerous challenges remain in realizing machines capable of autonomously conducting experiments. Coley et al. delve into these challenges in the automation of experimental and computational validations and the selection of experiments, while referring to studies on automated verification \cite{coley2020autonomousII}. They also point out the significance of removing hardware constraints to increase the automatability of research projects and reducing the costs associated with research automation \cite{coley2020autonomousII}. Zenil et al. also discuss the challenges of automating experiments and propose specific action plans \cite{zenil2023}.

Among these challenges, the development of robots capable of manipulating low-level actions as humans do, which is necessary to achieve a versatile automated experimental machine adaptable to diverse research tasks, seems to be exceedingly challenging. How to address these challenges will require further discussion.

\section{Additional Topics}

\subsection{Combining Question Formulation, Hypothesis Generation, and Hypothesis Verification}
Reflecting retrospectively on completed studies, it becomes evident that each study possesses its unique question, an accompanying hypothesis, and a process for verifying that hypothesis. Viewed from this perspective, research can be considered an endeavor that involves a sequence of constructing questions, generating hypotheses, and verifying these hypotheses, as classically described.

However, as we know, actual research is a highly complex, cyclical process of trial and error. Rarely do these tasks unfold as initially planned or occur just once in a single study. In practice, for instance, numerous questions and hypotheses might be generated even when formulating a single hypothesis, and not all of these lead to the final research outcome.\footnote{
The trial-and-error nature of activities is particularly significant in the context of discovery \cite{yanai2020hypothesis}.
} You will notice that even some major scientific discoveries throughout history were also made through these trials and errors \cite{hanson1965patterns,gribbin2022origin,whiteside1970before}.

Therefore, it is more accurate to view the construction of questions, the generation of hypotheses, and the verification of hypotheses as fundamental units for reducing uncertainty. In the research process, they are combined to gradually reduce the vast uncertainty inherent in the research endeavor. AI capable of conducting research is expected to master these flexible and complex operations. In this section, I will speculatively explore these characteristics of real research practice that have not been previously discussed

\subsubsection{Countless Questions, Hypotheses, and Verifications in a Single Research Process}
In the course of developing a single question or hypothesis, or in planning and preparing for a single verification, we generate countless questions and hypotheses, including implicit ones. Whether it's searching for problems, contemplating why a problem hasn't been solved, considering possible hypothesis candidates, planning verification, or doing anything else, we always pose questions and formulate hypotheses whenever dealing with unknowns or uncertainties. 

We also conduct a form of verification, whether implicit or explicit, and with varying degrees of simplicity, to generate plausible hypotheses. Generating plausible hypotheses requires having sufficient grounds to believe in their validity. These grounds could include knowledge from our memory, insights from recently researched literature, opinions from other researchers, or a belief in the simplicity of natural laws. Furthermore, we might conduct simple tests or even preliminary experiments to assess their plausibility. All these function as verification for researchers to be convinced.

Agents capable of conducting research should autonomously generate numerous questions and hypotheses as needed, and select the more plausible hypotheses through simple verifications during the knowledge production process. These are inevitable as long as uncertainties exist. How to realize such flexible agents remains an open question.

\subsubsection{Operations Apparently Unrelated to Knowledge Production}
\label{section-countless-seemingly-unrelated-operations-to-knowledge-production}

Research comprises numerous operations that may initially seem unrelated to knowledge production.\footnote{
Latour's anthropological study of daily practices in laboratories aptly illustrates these realities \cite{latour1987science}.
} In Section \ref{section-experimentation}, I argue that such tasks are essential in the context of experimentation. Most researchers would concur that daily academic activities are primarily characterized by these operations.

The construction of questions, generation of hypotheses, and verification of these hypotheses represent the core aims and functions in the knowledge production process. To implement these functions, performing operations like those mentioned above and combining them effectively to achieve desired objectives is crucial. Appropriately integrating these varied operations poses a significant challenge, even when tailored to a specific research question \cite{coley2020autonomousII}. 
To develop an agent capable of autonomously executing the entire research process, starting from question generation, replicating the flexibility of human action is indispensable.

\subsubsection{Discovering New Questions}

Researchers often begin with a specific question, only to discover an entirely unrelated question during their investigation. This new question, divergent from the original and its underlying purpose, can lead to a shift in research focus and potentially significant scientific breakthroughs. Given the inherent unpredictability in research, such discovery and redirection of focus are not rare phenomena.

If an agent designed for conducting research were tasked with a singular objective, such serendipitous discoveries might be overlooked. This is because the agent, focused on its predefined goal, may disregard new questions not aligned directly with its initial objective, regardless of their scientific value. To facilitate the agent's identification of such unrelated questions, it may be necessary to assign multiple objectives or a broader, overarching goal that accommodates both the original and emergent questions.

On the other hand, having a common high-level goal alone is not sufficient. For an agent to transition from its current question to a newly discovered one, it must be capable of evaluating which question is more valuable. The decision-making process regarding the value of a question, as discussed in Section \ref{section-deciding-what-knowledge-to-seek}, should encompass comparing multiple questions that share higher-order objectives. The development of such evaluative capabilities remains a challenging and open question.

\subsubsection{Incorporating Feedback from Verification Result}

In research, it is uncommon that the initial hypothesis is the answer of the posed question. Typically, research entails revising the hypothesis based on verification results and conducting subsequent rounds of testing. This iterative cycle of hypothesis revision and retesting is critical for scientific discovery. Therefore, an agent designed for conducting research should possess the ability to revise its hypotheses based on these outcomes of verification.

Efforts to automate incorporating feedback from verification results include studies in closed-cycle laboratory automation, as discussed in Section \ref{section-hypothesis-verification} and automation with scientific workflow \cite{gil2022will}. These have significantly contributed to automating the hypothesis revision cycle.

Despite these advancements, challenges persist in developing machines that autonomously analyze and respond to verification results like humans. When verification results are negative, pinpointing the exact cause is complex. This complexity arises because verification relies on a web of implicit and explicit hypotheses, any of which could contribute to the result \cite{sep-scientific-underdetermination}. The cause might be the primary hypothesis, underlying premises, auxiliary hypotheses, observations, experimental instruments, or a combination of these. A research-conducting agent must be capable of discerning the likely cause among these numerous candidates. Although it seems that humans do this well \cite{ren2023autonomous}, it is a challenging task for machines to do this autonomously.

Moreover, appropriate interpretation of experimental data is essential. Interpretations can vary based on the researcher's beliefs, prior knowledge, theoretical framework, and expectations \cite{hanson1965patterns}. Thus, verification results may undergo multiple reinterpretations, each potentially altering the hypothesis in need of revision. Additionally, as previously mentioned, researchers sometimes derive entirely different questions from these results and may temporarily halt their research. Current machines have yet to match this level of complex interpretation and adaptability in handling verification results that humans exhibit. An ideal autonomous research agent would be expected to possess these capabilities.

\subsection{Common Topics}
In the preceding sections, I have speculatively examined various key elements integral to research and their interplay. In this section, I aim to delve into topics that are universally relevant to these elements or that have not yet been explored in this paper. 

\subsubsection{Language Models}
Recent years have witnessed rapid advancements in language models \cite{zhao2023survey}, opening new frontiers in research. The future of research agents is undoubtedly intertwined with the insights provided by these models. This section explores various initiatives investigating the potential of language models in research.

Beginning with the transformative impact of the Transformer \cite{vaswani2017attention} and BERT \cite{devlin2018bert}, the concept of ``scaling law'' \cite{kaplan2020scaling} and ``foundation model'' \cite{bommasani2021opportunities} has catalyzed the development of large scale models pre-trained on extensive corpora. This has led to the creation of scientific language models like SciBERT \cite{beltagy2019scibert} and others \cite{cohan2020specter,singh2022scirepeval,nadkarni2021scientific,gupta2022matscibert,taylor2022galactica,azerbayev2023llemma,xie2023darwin,luo2022biogpt,yang2022gatortron,deng2023learning}. Given that scientific data is multimodal, attempts are emerging to construct general-purpose models using multimodal data as well \cite{li2023llava,tu2023towards,takeda2023foundation,nguyen2023climax}.

The development of GPTs, including GPT-3 \cite{brown2020language}, InstructGPT \cite{ouyang2022training}, and GPT-4 \cite{GPT4} and the advent of the web application called ChatGPT  \cite{ChatGPT} marked significant milestones. Their ability to perform various intellectual tasks has spurred research into their scientific applications. Emerging research examines the potential of LLMs, especially these GPTs, in various research fields, including the natural sciences \cite{ai4science2023impact,boiko2023emergent,qin2023gpt,bran2023chemcrow,white2022large,hatakeyama2023prompt,jablonka202314,guo2023can,zheng2023large,qian2023can,wysocka2023large,nori2023capabilities,wang2023large,singhal2023large}, mathematics and engineering \cite{bordt2023chatgpt,wu2023empirical,pursnani2023performance,zheng2023can,zhang2023automl,vijay2023prompt}, and social sciences \cite{koneru2023can,wang2023survey,bail2023can,ziems2023can,park2023generative,horton2023large,korinek2023generative,aher2023using}.

Additionally, there are efforts applicable to all research fields, which involve using GPTs for processing academic documents \cite{alzaabi2023chatgpt}. Some of them include paper search and reading \cite{elicit,scispace}, paper writing \cite{transformer2022can}, abstract generation \cite{gao2023comparing}, literature review generation \cite{aydin2022openai}, and peer review \cite{wexin2023can,liu2023reviewergpt,robertson2023gpt4}.\footnote{
Hosseini and Horbach discuss the influence of LLMs on the role of peer review \cite{hosseini2023fighting}.
}

The scope of applications of LLMs ranges from generating hypotheses to generating research questions finding research challenges \cite{liu2023creative,oppenlaender2023mapping,lahat2023evaluating}. Some studies even attempted to automate experimentation \cite{boiko2023emergent,qin2023gpt}. 

It is important to recognize that research and development in the field of language models are advancing rapidly. The potential applications for automating research are vast and constantly evolving, warranting further exploration and critical assessment of their impact on the scientific method.

\subsubsection{Incorporating Scientific Knowledge}

While discussing the challenges of enabling machines to conduct research, it's pertinent to acknowledge that even humans do not embark on research from a completely zero starting point. Firstly, humans are inherently equipped with brains and bodies evolved and developed for interpreting the world. Moreover, before engaging in research, we study the fundamental knowledge of the research fields. Thus, it would be reasonable to assume that agents should also possess basic knowledge before they begin conducting research. The advancement of language models has brought a wealth of knowledge to machines, but it is still considered necessary for them to acquire the knowledge required for research.

This is embodied in the concept of \textit{physics-informed machine learning} \cite{karniadakis2021physics}, where biases and scientific knowledge are integrated into AI to process scientific data. Karniadakis et al. \cite{karniadakis2021physics} and Hao et al. \cite{hao2022physics} provide a systematic overview of research in this domain. As previously discussed, imparting scientific knowledge through training on textual and multi-modal data is also a prevalent strategy.

Furthermore, just as humans continuously update their scientific knowledge, it's imperative for machines to not only embed knowledge through pre-training and inductive biases or retrieval during inference but also to continually update this knowledge. Specifically, it is crucial to acknowledge that knowledge produced by research is perpetually evolving. Thus, an agent must assimilate new knowledge, retain and update existing knowledge, and adapt to revisions in previously learned concepts. While the methodology for achieving this remains an open question, it is a subject of increasing debate \cite{kitano2021nobel,zenil2023future}.

\subsubsection{Autonomy, Generality, and Open-Endedness}

As previously emphasized, ongoing efforts are being made to enable machines to autonomously generate research questions, formulate hypotheses, and validate these hypotheses. However, the significant challenge remains in achieving this autonomy with minimal human intervention. Even in the realm of closed-loop research automation, which represents a substantial stride towards autonomy, full automation of all research processes is still an unrealized goal \cite{zenil2023,coley2020autonomous,coley2020autonomousII}.

This problem becomes particularly serious when attempting to enable machines to independently formulate research objectives, problems, and questions. If machines are to autonomously generate goals and questions, they must also be capable of independently generating and validating corresponding hypotheses. This necessitates a versatile approach to hypothesis generation and validation, adaptable to a wide range of questions.

In such scenarios, humans cannot provide predefined methods, potential hypotheses, or necessary information. Consequently, machines must be equipped to extract pertinent information from an open-ended environment, mirroring the human approach. Research, in essence, is a process of seeking information from the vast outer world of scientific data and processing it within the agent's cognitive framework, as highlighted in \cite{hope2022computational}. Assuming an open-ended environment means minimizing human-imposed constraints on this outer world and allowing the agent maximum freedom in selecting and processing information from the outer world.

Given these considerations, the extent to which autonomy should be expected from machines and the level of constraints that can be imposed without stifling their potential for autonomous hypothesis generation and validation, remains a critical and open question.

\subsubsection{Scientific Understanding}
While the primary focus of this paper has been on knowledge discovery, it's crucial to also consider another important goal of research: understanding. As Krenn et al. highlight, scientific discoveries can be made without understanding \cite{krenn2022scientific}, suggesting that facilitating scientific understanding in humans by automated machines demands more than the things discussed so far.

Scientific understanding in humans involves comprehending theories or hypotheses – both their nature and their underlying rationale. Therefore, an additional requirement seems necessary in the context of hypothesis generation. It remains unclear whether this pertains to the representation of generated hypotheses, the description of their generation process, or anything else. Identifying what is needed to be added in this process to foster scientific understanding, and how to implement it, is a significant issue.

Krenn et al. propose two conditions for an AI to achieve scientific understanding: 1. the ability to ``recognize qualitatively characteristic consequences of a theory without performing exact computations and use them in a new context'', and 2. the capacity to ``transfer its understanding to a human expert'' \cite{krenn2022scientific}.  As previously mentioned, the belief systems of humans and AI may differ. Thus whether AI having scientific understanding is necessary for bringing new scientific understanding to humans is unclear. Nonetheless, the capability to communicate understanding to humans is essential for bringing scientific understanding to humans. The explanation of machine prediction results has already been extensively researched and discussed as explainable AI \cite{arrieta2020explainable}, and its importance has already been widely pointed out in AI for Science research, so I will not delve further into it here.\footnote{
In considering how machine-driven scientific discoveries can facilitate human understanding, it might be worth exploring not only the enhancement of machine capabilities but also the expansion of human cognitive boundaries. While delving into this might border on speculative, it's a discussion that could yield valuable insights in the scientific community.
}

\subsubsection{Alignment}
Alignment is a critical concern in the development of autonomous AI researchers, akin to other areas of AI research. The primary concern is ensuring that these autonomous agents do not harm humans, a priority that becomes increasingly significant as we seek greater autonomy in machines. Given that knowledge is inherently value-neutral and can be used for benevolent or malevolent purposes, addressing this challenge is complex and necessitates ongoing discourse.

As previously discussed in Section \ref{section-question-construction}, alignment with human values and worldviews is crucial not only for safety but also for the relevance and effectiveness of AI-generated knowledge for humans. AI agents need to make value judgments aligned with human assessments of question quality and discern what is unknown and comprehensible from a human perspective, not just from their own standpoint.

These value judgments are often not explicitly stated in human-generated texts, indicating a need for proactive teaching of these values to AI. The methodology for implementing such teaching and ensuring alignment in a broader sense remains a complex issue that warrants further discussion and exploration.

This topic's complexity is amplified by the varying and evolving nature of human values and worldviews. As AI continues to advance, the challenge of continuously adapting these systems to align with human ethics and understanding becomes more pronounced. Future discussions should focus on developing robust frameworks and methodologies to achieve and maintain this alignment, ensuring that autonomous AI researchers contribute positively and safely to the scientific community.

\section{Ideas for Prototyping}
Realizing an autonomous intelligent agent capable of conducting research is an exceptionally challenging goal, one that will likely require a significant amount of time to achieve. The challenges discussed so far represent merely the tip of the iceberg found in speculative discussions; undoubtedly, many more critical issues remain unidentified. Therefore, it is crucial to begin by identifying these unknown challenges. A practical starting point might be the development of a simplified prototype of a research-capable agent. Such a prototype would allow us to explore and understand the challenges inherent to our goal during the prototyping process. In this section, I aim to discuss, in a speculative and brief manner, what might constitute such prototyping.

\subsection{Prototyping Agents that Conduct Research}

\subsubsection{Requirements for Prototype}
As discussed in Section \ref{section-question-hypothesis-verification}, Research seems to involve constructing questions, generating hypotheses, and verifying these hypotheses. Therefore, it appears appropriate for this prototype to incorporate these functions as distinct modules. The ``question construction'' module should take any input and formulate a question. The ``hypothesis generation'' module would then take this question as input and generate a hypothesis. Subsequently, the ``hypothesis verification'' module would take the hypothesis and provide verification results. By flexibly combining these modules at various levels, the prototype could mimic the research process.

For the prototype agent to function autonomously, human involvement in its design, implementation, and intervention should be minimized. Consequently, each module should autonomously gather information from the open-ended world, similar to how humans acquire information for research, while requiring minimal inputs. This means the agent should interact with the physical or digital realms to gather necessary information for research.

Moreover, to ensure the system's generality, the internal workings of each module should not overly rely on specific research topics. For instance, a verification method like experimentation tailored for specific physics research would not be applicable to psychological research. The human-designed elements of each module should be minimal, confined to what is essential for the module's function.

Creating a system that simultaneously meets the criteria of autonomy and generality, while effectively constructing questions, generating hypotheses, and verifying them within this abstract framework, is impractical, even in simpler scenarios. Thus, it may be necessary to introduce some constraints to this abstract framework. Discussing the nature, necessity, and potential relaxation of these constraints could shed light on the challenges in realizing an autonomous research agent. To initiate this discussion, I will present some candidates for potential constraints.

\subsubsection{Candidate Constraints in Prototyping}

In Section \ref{section-what-is-research}, I discussed the perspective of research as a process of updating beliefs and the potential for autonomously constructing verification from foundational concepts, as well as autonomously assessing the value of questions. However, these concepts are visionary and present significant challenges, making it unrealistic to expect immediate, meaningful outcomes for humans through prototyping. Therefore, it would be beneficial to start by prototyping agents that can master existing human values and research methods.

As mentioned in Section \ref{section-question-construction}, formulating questions from open-ended situations is an exceptionally challenging task, often with even no clear starting point. A pragmatic approach would be to predetermine the inputs for question construction, rather than relying on unrestricted information sources. A viable input could be a high-level goal, commonly assumed in many studies. Specifically, it would be beneficial to start with high-level goals recognized as research objectives in specific research fields.

A significant obstacle in developing a fully autonomous research agent, as discussed in Section \ref{section-countless-seemingly-unrelated-operations-to-knowledge-production}, is the need for expertise in complex low-level actions. Developing a robot capable of free physical world interaction like humans remains a formidable challenge. Hence, for prototyping, focusing initially on research confined to computational environments appears more manageable. While creating an agent capable of operating freely within a computer environment is also challenging, it is arguably more feasible than one operating in the physical world. Indeed, there have been efforts to enable language models to perform various computer operations \cite{openinterpreter,openai_chatgpt_plugins_code_interpreter_2023}, and to operate web browsers \cite{nakano2021webgpt,act1}.

The primary objective of prototyping is to concretize a concept, however rudimentary, and to identify challenges. Thus, it seems prudent to initially limit the prototype's environment to the digital realm, while waiting for advancements in foundational research that could enable free activity in the physical world.

The ideas presented here are merely initial suggestions and are neither definitive nor exhaustive. In the prototyping phase, it is crucial to discuss the extent and nature of the constraints to be applied. More suitable constraints are likely to emerge as these discussions evolve.

\subsubsection{Implementing Each Module with Large Language Models}

Given the need for generality and considering the remarkable capabilities of LLMs, it seems inevitable that each module in prototypes would be instantiated as an LLM. As highlighted in previous sections, there are emerging studies focused on constructing automated research pipelines using LLMs. I propose that our initial prototyping efforts should focus on creating autonomous research agents modeled on these LLM pipelines, in alignment with current endeavors to develop autonomous agents utilizing language models \cite{wang2023survey,xi2023rise}.

Here is a provisional concept, modeled after a typical autonomous agent. The research agent begins by formulating a question based on a high-level goal provided by a human. Following the posing of this question, the agent autonomously generates hypotheses to address it, and then proceeds to verify these hypotheses. After obtaining the final verification results, they are analyzed in relation to the initial objective and research question, prompting the generation of subsequent questions. This process – comprising question formulation, hypothesis generation, and hypothesis verification – is iteratively and hierarchically repeated to conduct research.

The agent is envisioned to perform four fundamental actions: 1) Formulating questions, 2) Determining task completion, 3) Verifying hypotheses, and 4) Executing low-level computer operations. The processes of question formulation, hypothesis generation, and verification primarily involve executing these low-level computer operations.

When the agent opts to generate a question, it temporarily pauses its current task, such as hypothesis verification, and initiates hypothesis generation for the new question. Upon completing hypothesis generation, the agent decides whether to proceed with verification. Following verification, it updates the hypotheses based on the results. Whether or not the hypotheses are verified, the agent then resumes the higher-level process that was previously paused, incorporating the results of the low-level process. In this way, the agent continuously cycles through lower-level tasks of question construction, hypothesis generation, and verification until the highest-level hypothesis is formulated. When the highest-level hypothesis – the response to the original question – is ready, the agent always proceeds to its verification.

To ensure the system's adaptability to a wide range of research questions, the prompts given to the LLMs should be composed of only general instructions. For example, an instruction like ``generate a hypothesis for the following question'' is sufficiently generic to apply to any research question. However, providing such instructions alone is unlikely to spontaneously yield research outcomes, so there may be a need for additional auxiliary instructions that are as general as possible; identifying these is one of the main goals of prototyping.

For open-ended operations within a computer environment, ideally, the LLMs should have access to nearly all operations on the computer. As previously mentioned, initiatives to develop language models capable of executing any action in such environments are underway \cite{openai_chatgpt_plugins_code_interpreter_2023,openinterpreter}. Minimal access to web browsers, search engines, or shells may be permissible, but reliance on custom corpora or predefined hypothesis spaces should be avoided. Successful autonomous research under these conditions would indeed demonstrate the system's capacity for independent research.

\subsubsection{Agents that Conduct Machine Learning Research}
To effectively provide a high-level goal for the prototype, it is necessary to select objectives from a specific research field that align with the constraints previously outlined and are conducive to prototyping.

I propose that machine learning research is an ideal candidate for such prototyping. First, many aspects of machine learning research, including verification, can be conducted entirely on a computer, thus meeting the constraints we have established. Second, the field typically features shorter research cycles compared to other disciplines, which allows for more rapid feedback for the prototype. Third, machine learning not only forms a foundational technology across various research fields but is crucial for developing a research-capable agent itself. Automating machine learning research would thus not only contribute to the automation of research processes in numerous other areas but also advance our primary objective. Finally, there already have been significant efforts towards automation in machine learning, such as AutoML \cite{hutter2019automated,bischl2023hyperparameter,lindauer2020best,white2023neural} and MLOps \cite{kreuzberger2023machine}. Particularly in recent years, there have been attempts to utilize language models for these tasks \cite{zheng2023can,zhang2023automl,vijay2023prompt}. These existing efforts are likely to provide valuable support in developing the prototype.

In conclusion, initiating the prototyping of autonomous agents, composed of language models and given the most general instructions possible, and focusing on specific types of machine learning research appears to be a strategic choice. While such efforts are already underway, I anticipate that increased participation in this area will significantly accelerate this movement.

\subsection{Prototyping Agents that Conduct Peer Review}
To identify challenges associated with creating agents capable of conducting research, another promising initial step could be to target the automation of the academic peer review process. This approach presents several advantages, which I will detail in the following section.

\subsubsection{Why Aim for Agents Capable of Conducting Peer Review?}

Firstly, the competencies necessary for peer review closely align with those required by a research-capable agent. This similarity arises because peer review fundamentally involves evaluating critical aspects of research, such as the soundness of the verification.

Secondly, automating peer review might present fewer challenges compared to developing a fully autonomous research agent. The distinction lies in the scope of tasks: peer review primarily entails assessing whether research incorporates the necessary elements, whereas a research-capable agent must not only evaluate but also synthesize these elements. As a preliminary step in prototyping, addressing simpler problems like peer review could effectively highlight key challenges.

Thirdly, peer review is a universal practice across various research fields. Insights gained from automating this process can thus contribute significantly to the development of a general research agent, applicable in multiple disciplines.

Fourthly, peer review predominantly involves textual analysis and does not require physical or extensive digital interactions, unlike conducting research autonomously. Although it may involve searches to review existing literature, tasks such as performing experiments are typically not necessary. Given the advancements in LLMs, we are now better equipped to handle complex textual tasks. This focus on text-based evaluation is beneficial for pinpointing specific challenges in achieving our broader objective.

Finally, peer review encompasses the evaluation of subjective aspects like the ``significance'' of a research question. As discussed in Section \ref{section-question-construction}, understanding how humans assess such value in research is crucial, especially considering that alignment with human values is a significant challenge. Peer review offers a unique opportunity to observe and analyze these value judgments explicitly. Therefore, beginning with the automation of peer reviews could provide valuable insights into human evaluative processes in research.

\subsubsection{Peer Review Automation}

A considerable body of research has been devoted to automating various aspects of the peer review process. Efforts have included automating the generation of reviews \cite{yuan2022can,yuan2022kid,wang2020reviewrobot}, screening papers \cite{schulz2022future}, assessing research papers \cite{kousha2022artificial}, and assigning reviewers \cite{zhao2022reviewer}, among other tasks. In line with trends across other fields, recent years have witnessed a surge in studies exploring the use of LLMs for automating peer review \cite{wexin2023can,liu2023reviewergpt,robertson2023gpt4,hosseini2023fighting}. For a more comprehensive understanding of traditional research in this area, Kousha et al. \cite{kousha2022artificial} and Lin et al. \cite{lin2021automated1} have conducted extensive literature reviews.

Considering the goals of prototyping, it is desirable that such efforts already exist. The insights and findings from these prior attempts would help further discussions on developing AI capable of conducting peer review. 

\section{Conclusion}
In this paper, I have undertaken a speculative exploration of the concept of an artificial agent capable of conducting research. The initial discussion centered on characterizing what constitutes research, tentatively framing it as the process of updating beliefs in hypotheses. Subsequently, I delved into the critical elements of research: the construction of questions, generation of hypotheses, and their verification. This paper then briefly looks at the common themes to these elements. Following the discussions, I highlighted the significance of identifying challenges in realizing such agents and proposed preliminary ideas for prototyping.

It is important to acknowledge that the discussions in this paper are purely speculative. The definition of research provided is provisional, the challenges and implications discussed represent only a fraction of the myriad possibilities, and the ideas for prototyping are rudimentary, akin to early-stage experiments. Furthermore, the literature referenced is not exhaustive, omitting many pivotal works. My ability to evaluate each reference thoroughly may have been limited, potentially leading to partial perspectives or inaccuracies. I plan to update this paper in the future to address these limitations. I greatly value feedback and corrections from readers, as they will be crucial in improving this work.

The primary motivation for publishing this paper in its current nascent form, despite its  numerous limitations, is to serve as an initial step towards future exploration into the concept of a research-capable artificial agent. In order to realize agents capable of conducting research, there must still be many issues that need to be discussed. I hope that the discussion surrounding this concept will become more active to accelerate the development of research-capable agents.

\bibliographystyle{unsrt}
\bibliography{ref}

\begin{thebibliography}{100}

\bibitem{zenil2023future}
Hector Zenil, Jesper Tegn{\'e}r, Felipe~S Abrah{\~a}o, Alexander Lavin, Vipin
  Kumar, Jeremy~G Frey, Adrian Weller, Larisa Soldatova, Alan~R Bundy,
  Nicholas~R Jennings, et~al.
\newblock The future of fundamental science led by generative closed-loop
  artificial intelligence.
\newblock {\em arXiv preprint arXiv:2307.07522}, 2023.

\bibitem{kitano2021nobel}
Hiroaki Kitano.
\newblock Nobel turing challenge: creating the engine for scientific discovery.
\newblock {\em npj Systems Biology and Applications}, 7(1):29, 2021.

\bibitem{lindsay1993dendral}
Robert~K Lindsay, Bruce~G Buchanan, Edward~A Feigenbaum, and Joshua Lederberg.
\newblock Dendral: a case study of the first expert system for scientific
  hypothesis formation.
\newblock {\em Artificial intelligence}, 61(2):209--261, 1993.

\bibitem{langley1987scientific}
Pat Langley.
\newblock {\em Scientific discovery: Computational explorations of the creative
  processes}.
\newblock MIT press, 1987.

\bibitem{king2004functional}
Ross~D King, Kenneth~E Whelan, Ffion~M Jones, Philip~GK Reiser, Christopher~H
  Bryant, Stephen~H Muggleton, Douglas~B Kell, and Stephen~G Oliver.
\newblock Functional genomic hypothesis generation and experimentation by a
  robot scientist.
\newblock {\em Nature}, 427(6971):247--252, 2004.

\bibitem{zenil2023}
Hector Zenil and Ross~D. King.
\newblock {\em The Automated AI-driven Future of Scientific Discovery}, pages
  679--691.
\newblock World Scientific, 2023.

\bibitem{wang2023scientific}
Hanchen Wang, Tianfan Fu, Yuanqi Du, Wenhao Gao, Kexin Huang, Ziming Liu, Payal
  Chandak, Shengchao Liu, Peter Van~Katwyk, Andreea Deac, et~al.
\newblock Scientific discovery in the age of artificial intelligence.
\newblock {\em Nature}, 620(7972):47--60, 2023.

\bibitem{xu2021artificial}
Yongjun Xu, Xin Liu, Xin Cao, Changping Huang, Enke Liu, Sen Qian, Xingchen
  Liu, Yanjun Wu, Fengliang Dong, Cheng-Wei Qiu, et~al.
\newblock Artificial intelligence: A powerful paradigm for scientific research.
\newblock {\em The Innovation}, 2(4):100179, 2021.

\bibitem{zhang2023artificial}
Xuan Zhang, Limei Wang, Jacob Helwig, Youzhi Luo, Cong Fu, Yaochen Xie, Meng
  Liu, Yuchao Lin, Zhao Xu, Keqiang Yan, et~al.
\newblock Artificial intelligence for science in quantum, atomistic, and
  continuum systems.
\newblock {\em arXiv preprint arXiv:2307.08423}, 2023.

\bibitem{ai4science2023impact}
Microsoft~Research AI4Science and Microsoft~Azure Quantum.
\newblock The impact of large language models on scientific discovery: a
  preliminary study using gpt-4.
\newblock {\em arXiv preprint arXiv:2311.07361}, 2023.

\bibitem{coley2020autonomousII}
Connor~W Coley, Natalie~S Eyke, and Klavs~F Jensen.
\newblock Autonomous discovery in the chemical sciences part ii: outlook.
\newblock {\em Angewandte Chemie International Edition}, 59(52):23414--23436,
  2020.

\bibitem{coley2020autonomous}
Connor~W Coley, Natalie~S Eyke, and Klavs~F Jensen.
\newblock Autonomous discovery in the chemical sciences part i: Progress.
\newblock {\em Angewandte Chemie International Edition}, 59(51):22858--22893,
  2020.

\bibitem{hope2022computational}
Tom Hope, Doug Downey, Oren Etzioni, Daniel~S Weld, and Eric Horvitz.
\newblock A computational inflection for scientific discovery.
\newblock {\em arXiv preprint arXiv:2205.02007}, 2022.

\bibitem{national2022automated}
National~Academies of~Sciences~Engineering, Medicine, et~al.
\newblock {\em Automated Research Workflows for Accelerated Discovery: Closing
  the Knowledge Discovery Loop}.
\newblock The National Academies Press, 2022.

\bibitem{chalmers2013thing}
Alan~F Chalmers.
\newblock {\em What is this thing called science?}
\newblock Hackett Publishing, 2013.

\bibitem{sep-scientific-method}
Brian Hepburn and Hanne Andersen.
\newblock {Scientific Method}.
\newblock In Edward~N. Zalta, editor, {\em The {Stanford} Encyclopedia of
  Philosophy}. Metaphysics Research Lab, Stanford University, {S}ummer 2021
  edition, 2021.

\bibitem{booth2003craft}
Wayne~C Booth, Gregory~G Colomb, and Joseph~M Williams.
\newblock {\em The craft of research}.
\newblock University of Chicago press, 2003.

\bibitem{manual2015guidelines}
Frascati Manual et~al.
\newblock Guidelines for collecting and reporting data on research and
  experimental development.
\newblock {\em URL: http://www. oecd.
  org/sti/frascati-manual-2015-9789264239012-en. htm}, 2015.

\bibitem{balzandistributed}
Francesco Balzan-francesco, John Campbell, Karl Friston, Maxwell~James
  Ramstead, Daniel Friedman, and Axel Constant.
\newblock Distributed science-the scientific process as multi-scale active
  inference.
\newblock {\em OSF Preprints}, 2023.

\bibitem{sep-epistemology}
Matthias Steup and Ram Neta.
\newblock {Epistemology}.
\newblock In Edward~N. Zalta, editor, {\em The {Stanford} Encyclopedia of
  Philosophy}. Metaphysics Research Lab, Stanford University, {F}all 2020
  edition, 2020.

\bibitem{gettier1963justified}
Edmund~L Gettier.
\newblock Is justified true belief knowledge?
\newblock {\em analysis}, 23(6):121--123, 1963.

\bibitem{otsuka2022thinking}
Jun Otsuka.
\newblock {\em Thinking About Statistics: The Philosophical Foundations}.
\newblock Taylor \& Francis, 2022.

\bibitem{lahav2022search}
Dan Lahav, Jon~Saad Falcon, Bailey Kuehl, Sophie Johnson, Sravanthi Parasa,
  Noam Shomron, Duen~Horng Chau, Diyi Yang, Eric Horvitz, Daniel~S Weld, et~al.
\newblock A search engine for discovery of scientific challenges and
  directions.
\newblock In {\em Proceedings of the AAAI Conference on Artificial
  Intelligence}, volume~36, pages 11982--11990, 2022.

\bibitem{oppenlaender2023mapping}
Jonas Oppenlaender and Joonas H{\"a}m{\"a}l{\"a}inen.
\newblock Mapping the challenges of hci: An application and evaluation of
  chatgpt and gpt-4 for cost-efficient question answering.
\newblock {\em arXiv preprint arXiv:2306.05036}, 2023.

\bibitem{surita2020can}
Gabriela Surita, Rodrigo Nogueira, and Roberto Lotufo.
\newblock Can questions summarize a corpus? using question generation for
  characterizing covid-19 research.
\newblock {\em arXiv preprint arXiv:2009.09290}, 2020.

\bibitem{wang2019paperrobot}
Qingyun Wang, Lifu Huang, Zhiying Jiang, Kevin Knight, Heng Ji, Mohit Bansal,
  and Yi~Luan.
\newblock Paperrobot: Incremental draft generation of scientific ideas.
\newblock {\em arXiv preprint arXiv:1905.07870}, 2019.

\bibitem{krenn2020predicting}
Mario Krenn and Anton Zeilinger.
\newblock Predicting research trends with semantic and neural networks with an
  application in quantum physics.
\newblock {\em Proceedings of the National Academy of Sciences},
  117(4):1910--1916, 2020.

\bibitem{krenn2022predicting}
Mario Krenn, Lorenzo Buffoni, Bruno Coutinho, Sagi Eppel, Jacob~Gates Foster,
  Andrew Gritsevskiy, Harlin Lee, Yichao Lu, Joao~P Moutinho, Nima Sanjabi,
  et~al.
\newblock Predicting the future of ai with ai: High-quality link prediction in
  an exponentially growing knowledge network.
\newblock {\em arXiv preprint arXiv:2210.00881}, 2022.

\bibitem{pan2019recent}
Liangming Pan, Wenqiang Lei, Tat-Seng Chua, and Min-Yen Kan.
\newblock Recent advances in neural question generation.
\newblock {\em arXiv preprint arXiv:1905.08949}, 2019.

\bibitem{zhang2021review}
Ruqing Zhang, Jiafeng Guo, Lu~Chen, Yixing Fan, and Xueqi Cheng.
\newblock A review on question generation from natural language text.
\newblock {\em ACM Transactions on Information Systems (TOIS)}, 40(1):1--43,
  2021.

\bibitem{schmidhuber1991possibility}
J{\"u}rgen Schmidhuber.
\newblock A possibility for implementing curiosity and boredom in
  model-building neural controllers.
\newblock In {\em Proc. of the international conference on simulation of
  adaptive behavior: From animals to animats}, pages 222--227, 1991.

\bibitem{liu2023creative}
Yiren Liu, Mengxia Yu, Meng Jiang, and Yun Huang.
\newblock Creative research question generation for human-computer interaction
  research.
\newblock In {\em Joint Proceedings of the ACM IUI Workshop}, 2023.

\bibitem{lahat2023evaluating}
Adi Lahat, Eyal Shachar, Benjamin Avidan, Zina Shatz, Benjamin~S Glicksberg,
  and Eyal Klang.
\newblock Evaluating the use of large language model in identifying top
  research questions in gastroenterology.
\newblock {\em Scientific reports}, 13(1):4164, 2023.

\bibitem{watson_2021}
Lani Watson.
\newblock What is a question.
\newblock {\em Royal Institute of Philosophy Supplements}, 89:273–297, 2021.

\bibitem{taylor1962process}
Robert~S Taylor.
\newblock The process of asking questions.
\newblock {\em American documentation}, 13(4):391--396, 1962.

\bibitem{wilson1997information}
Tom~D Wilson.
\newblock Information behaviour: an interdisciplinary perspective.
\newblock {\em Information processing \& management}, 33(4):551--572, 1997.

\bibitem{case2016looking}
Donald~O Case and Lisa~M Given.
\newblock {\em Looking for information: A survey of research on information
  seeking, needs, and behavior}.
\newblock Emerald Group Publishing, 2016.

\bibitem{kidd2015psychology}
Celeste Kidd and Benjamin~Y Hayden.
\newblock The psychology and neuroscience of curiosity.
\newblock {\em Neuron}, 88(3):449--460, 2015.

\bibitem{aubret2019survey}
Arthur Aubret, Laetitia Matignon, and Salima Hassas.
\newblock A survey on intrinsic motivation in reinforcement learning.
\newblock {\em arXiv preprint arXiv:1908.06976}, 2019.

\bibitem{alvesson2013constructing}
Mats Alvesson and Jorgen Sandberg.
\newblock {\em Constructing research questions: Doing interesting research}.
\newblock Sage, 2013.

\bibitem{hulley2007designing}
Stephen~B Hulley.
\newblock {\em Designing clinical research}.
\newblock Lippincott Williams \& Wilkins, 2007.

\bibitem{alon2009choose}
Uri Alon.
\newblock How to choose a good scientific problem.
\newblock {\em Molecular cell}, 35(6):726--728, 2009.

\bibitem{huntington2021effect}
Nick Huntington-Klein.
\newblock {\em The effect: An introduction to research design and causality}.
\newblock CRC Press, 2021.

\bibitem{oudeyer2018computational}
Pierre-Yves Oudeyer.
\newblock Computational theories of curiosity-driven learning.
\newblock {\em arXiv preprint arXiv:1802.10546}, 2018.

\bibitem{sep-scientific-discovery}
Jutta Schickore.
\newblock {Scientific Discovery}.
\newblock In Edward~N. Zalta and Uri Nodelman, editors, {\em The {Stanford}
  Encyclopedia of Philosophy}. Metaphysics Research Lab, Stanford University,
  {W}inter 2022 edition, 2022.

\bibitem{hanson1965patterns}
Norwood~Russell Hanson.
\newblock {\em Patterns of discovery: An inquiry into the conceptual
  foundations of science}.
\newblock CUP Archive, 1965.

\bibitem{magnani2011abduction}
Lorenzo Magnani.
\newblock {\em Abduction, reason and science: Processes of discovery and
  explanation}.
\newblock Springer Science \& Business Media, 2011.

\bibitem{gentner2002analogy}
Dedre Gentner.
\newblock Analogy in scientific discovery: The case of johannes kepler.
\newblock In {\em Model-based reasoning}, pages 21--39. Springer, 2002.

\bibitem{dasgupta2017hypotheses}
Ishita Dasgupta, Eric Schulz, and Samuel~J Gershman.
\newblock Where do hypotheses come from?
\newblock {\em Cognitive psychology}, 96:1--25, 2017.

\bibitem{jumper2021highly}
John Jumper, Richard Evans, Alexander Pritzel, Tim Green, Michael Figurnov,
  Olaf Ronneberger, Kathryn Tunyasuvunakool, Russ Bates, Augustin
  {\v{Z}}{\'\i}dek, Anna Potapenko, et~al.
\newblock Highly accurate protein structure prediction with alphafold.
\newblock {\em Nature}, 596(7873):583--589, 2021.

\bibitem{merchant2023scaling}
Amil Merchant, Simon Batzner, Samuel~S. Schoenholz, Muratahan Aykol, Gowoon
  Cheon, and Ekin~Dogus Cubuk.
\newblock Scaling deep learning for materials discovery.
\newblock {\em Nature}, 2023.

\bibitem{kramer2023automated}
Stefan Kramer, Mattia Cerrato, Sa{\v{s}}o D{\v{z}}eroski, and Ross King.
\newblock Automated scientific discovery: From equation discovery to autonomous
  discovery systems.
\newblock {\em arXiv preprint arXiv:2305.02251}, 2023.

\bibitem{kang2022augmenting}
Hyeonsu~B Kang, Xin Qian, Tom Hope, Dafna Shahaf, Joel Chan, and Aniket Kittur.
\newblock Augmenting scientific creativity with an analogical search engine.
\newblock {\em ACM Transactions on Computer-Human Interaction}, 2022.

\bibitem{chan2018solvent}
Joel Chan, Joseph~Chee Chang, Tom Hope, Dafna Shahaf, and Aniket Kittur.
\newblock Solvent: A mixed initiative system for finding analogies between
  research papers.
\newblock {\em Proceedings of the ACM on Human-Computer Interaction},
  2(CSCW):1--21, 2018.

\bibitem{wang2023learning}
Qingyun Wang, Doug Downey, Heng Ji, and Tom Hope.
\newblock Learning to generate novel scientific directions with contextualized
  literature-based discovery.
\newblock {\em arXiv preprint arXiv:2305.14259}, 2023.

\bibitem{xu2023exploring}
Yi~Xu, Shuqian Sheng, Bo~Xue, Luoyi Fu, Xinbing Wang, and Chenghu Zhou.
\newblock Exploring and verbalizing academic ideas by concept co-occurrence.
\newblock {\em arXiv preprint arXiv:2306.02282}, 2023.

\bibitem{yang2023large}
Zonglin Yang, Xinya Du, Junxian Li, Jie Zheng, Soujanya Poria, and Erik
  Cambria.
\newblock Large language models for automated open-domain scientific hypotheses
  discovery.
\newblock {\em arXiv preprint arXiv:2309.02726}, 2023.

\bibitem{park2023can}
Yang~Jeong Park, Daniel Kaplan, Zhichu Ren, Chia-Wei Hsu, Changhao Li, Haowei
  Xu, Sipei Li, and Ju~Li.
\newblock Can chatgpt be used to generate scientific hypotheses?
\newblock {\em arXiv preprint arXiv:2304.12208}, 2023.

\bibitem{guo2017calibration}
Chuan Guo, Geoff Pleiss, Yu~Sun, and Kilian~Q Weinberger.
\newblock On calibration of modern neural networks.
\newblock In {\em International conference on machine learning}, pages
  1321--1330. PMLR, 2017.

\bibitem{maynez2020faithfulness}
Joshua Maynez, Shashi Narayan, Bernd Bohnet, and Ryan McDonald.
\newblock On faithfulness and factuality in abstractive summarization.
\newblock {\em arXiv preprint arXiv:2005.00661}, 2020.

\bibitem{david2010history}
M~Burton David.
\newblock {\em The history of mathematics an introduction}.
\newblock McGraw-Hill Professional, 2010.

\bibitem{bochner1968role}
Salomon Bochner and Banesh Hoffmann.
\newblock The role of mathematics in the rise of science.
\newblock {\em American Journal of Physics}, 36(6):564--565, 1968.

\bibitem{heisenberg2008abstraction}
Werner Heisenberg.
\newblock Abstraction in modern science.
\newblock {\em Nishina Memorial Lectures}, pages 1--16, 2008.

\bibitem{rabe2021towards}
Markus~N Rabe and Christian Szegedy.
\newblock Towards the automatic mathematician.
\newblock In {\em Automated Deduction--CADE 28: 28th International Conference
  on Automated Deduction, Virtual Event, July 12--15, 2021, Proceedings 28},
  pages 25--37. Springer International Publishing, 2021.

\bibitem{imani2023mathprompter}
Shima Imani, Liang Du, and Harsh Shrivastava.
\newblock Mathprompter: Mathematical reasoning using large language models.
\newblock {\em arXiv preprint arXiv:2303.05398}, 2023.

\bibitem{goyal2022inductive}
Anirudh Goyal and Yoshua Bengio.
\newblock Inductive biases for deep learning of higher-level cognition.
\newblock {\em Proceedings of the Royal Society A}, 478(2266):20210068, 2022.

\bibitem{chaloner1995bayesian}
Kathryn Chaloner and Isabella Verdinelli.
\newblock Bayesian experimental design: A review.
\newblock {\em Statistical science}, pages 273--304, 1995.

\bibitem{baker2019basic}
N~Baker et~al.
\newblock Basic research needs workshop for scientific machine learning: Core
  technologies for artificial intelligence.
\newblock {\em Document prepared for Department of Energy Advanced Scientific
  Computing Research, USA}, 10, 2019.

\bibitem{wadden2020fact}
David Wadden, Shanchuan Lin, Kyle Lo, Lucy~Lu Wang, Madeleine van Zuylen, Arman
  Cohan, and Hannaneh Hajishirzi.
\newblock Fact or fiction: Verifying scientific claims.
\newblock {\em arXiv preprint arXiv:2004.14974}, 2020.

\bibitem{guo2022survey}
Zhijiang Guo, Michael Schlichtkrull, and Andreas Vlachos.
\newblock A survey on automated fact-checking.
\newblock {\em Transactions of the Association for Computational Linguistics},
  10:178--206, 2022.

\bibitem{koneru2023can}
Sai Koneru, Jian Wu, and Sarah Rajtmajer.
\newblock Can large language models discern evidence for scientific hypotheses?
  case studies in the social sciences.
\newblock {\em arXiv preprint arXiv:2309.06578}, 2023.

\bibitem{dhuliawala2023chain}
Shehzaad Dhuliawala, Mojtaba Komeili, Jing Xu, Roberta Raileanu, Xian Li, Asli
  Celikyilmaz, and Jason Weston.
\newblock Chain-of-verification reduces hallucination in large language models.
\newblock {\em arXiv preprint arXiv:2309.11495}, 2023.

\bibitem{kousha2022artificial}
Kayvan Kousha and Mike Thelwall.
\newblock Artificial intelligence technologies to support research assessment:
  A review.
\newblock {\em arXiv preprint arXiv:2212.06574}, 2022.

\bibitem{lin2021automated1}
Jialiang Lin, Jiaxin Song, Zhangping Zhou, and Xiaodong Shi.
\newblock Automated scholarly paper review: possibility and challenges.
\newblock {\em arXiv preprint arXiv:2111.07533}, 2021.

\bibitem{radder2009philosophy}
Hans Radder.
\newblock The philosophy of scientific experimentation: a review.
\newblock {\em Automated experimentation}, 1(1):1--8, 2009.

\bibitem{holland2020automation}
Ian Holland and Jamie~A Davies.
\newblock Automation in the life science research laboratory.
\newblock {\em Frontiers in Bioengineering and Biotechnology}, 8:571777, 2020.

\bibitem{abolhasani2023rise}
Milad Abolhasani and Eugenia Kumacheva.
\newblock The rise of self-driving labs in chemical and materials sciences.
\newblock {\em Nature Synthesis}, pages 1--10, 2023.

\bibitem{burger2020mobile}
Benjamin Burger, Phillip~M Maffettone, Vladimir~V Gusev, Catherine~M Aitchison,
  Yang Bai, Xiaoyan Wang, Xiaobo Li, Ben~M Alston, Buyi Li, Rob Clowes, et~al.
\newblock A mobile robotic chemist.
\newblock {\em Nature}, 583(7815):237--241, 2020.

\bibitem{yachie2017robotic}
Nozomu Yachie and Tohru Natsume.
\newblock Robotic crowd biology with maholo labdroids.
\newblock {\em Nature biotechnology}, 35(4):310--312, 2017.

\bibitem{boiko2023emergent}
Daniil~A Boiko, Robert MacKnight, and Gabe Gomes.
\newblock Emergent autonomous scientific research capabilities of large
  language models.
\newblock {\em arXiv preprint arXiv:2304.05332}, 2023.

\bibitem{qin2023gpt}
Xiaokai Qin, Mingda Song, Yangguan Chen, Zhehong Ai, and Jing Jiang.
\newblock Gpt-lab: Next generation of optimal chemistry discovery by gpt driven
  robotic lab.
\newblock {\em arXiv preprint arXiv:2309.16721}, 2023.

\bibitem{charness2023generation}
Gary Charness, Brian Jabarian, and John~A List.
\newblock Generation next: Experimentation with ai.
\newblock Technical report, National Bureau of Economic Research, 2023.

\bibitem{hao2022physics}
Zhongkai Hao, Songming Liu, Yichi Zhang, Chengyang Ying, Yao Feng, Hang Su, and
  Jun Zhu.
\newblock Physics-informed machine learning: A survey on problems, methods and
  applications.
\newblock {\em arXiv preprint arXiv:2211.08064}, 2022.

\bibitem{karniadakis2021physics}
George~Em Karniadakis, Ioannis~G Kevrekidis, Lu~Lu, Paris Perdikaris, Sifan
  Wang, and Liu Yang.
\newblock Physics-informed machine learning.
\newblock {\em Nature Reviews Physics}, 3(6):422--440, 2021.

\bibitem{yanai2020hypothesis}
Itai Yanai and Martin Lercher.
\newblock A hypothesis is a liability, 2020.

\bibitem{gribbin2022origin}
John Gribbin and Mary Gribbin.
\newblock {\em On The Origin of Evolution: Tracing ‘Darwin’s Dangerous
  Idea’from Aristotle to DNA}.
\newblock Rowman \& Littlefield, 2022.

\bibitem{whiteside1970before}
Derek~Thomas Whiteside.
\newblock Before the principia: The maturing of newton's thoughts on dynamical
  astronomy, 1664--1684.
\newblock {\em Journal for the History of Astronomy}, 1(1):5--19, 1970.

\bibitem{latour1987science}
Bruno Latour.
\newblock {\em Science in action: How to follow scientists and engineers
  through society}.
\newblock Harvard university press, 1987.

\bibitem{gil2022will}
Yolanda Gil.
\newblock Will ai write scientific papers in the future?
\newblock {\em AI Magazine}, 42(4):3--15, 2022.

\bibitem{sep-scientific-underdetermination}
Kyle Stanford.
\newblock {Underdetermination of Scientific Theory}.
\newblock In Edward~N. Zalta and Uri Nodelman, editors, {\em The {Stanford}
  Encyclopedia of Philosophy}. Metaphysics Research Lab, Stanford University,
  {S}ummer 2023 edition, 2023.

\bibitem{ren2023autonomous}
Zhichu Ren, Zekun Ren, Zhen Zhang, Tonio Buonassisi, and Ju~Li.
\newblock Autonomous experiments using active learning and ai.
\newblock {\em Nature Reviews Materials}, 8(9):563--564, 2023.

\bibitem{zhao2023survey}
Wayne~Xin Zhao, Kun Zhou, Junyi Li, Tianyi Tang, Xiaolei Wang, Yupeng Hou,
  Yingqian Min, Beichen Zhang, Junjie Zhang, Zican Dong, et~al.
\newblock A survey of large language models.
\newblock {\em arXiv preprint arXiv:2303.18223}, 2023.

\bibitem{vaswani2017attention}
Ashish Vaswani, Noam Shazeer, Niki Parmar, Jakob Uszkoreit, Llion Jones,
  Aidan~N Gomez, {\L}ukasz Kaiser, and Illia Polosukhin.
\newblock Attention is all you need.
\newblock {\em Advances in neural information processing systems}, 30, 2017.

\bibitem{devlin2018bert}
Jacob Devlin, Ming-Wei Chang, Kenton Lee, and Kristina Toutanova.
\newblock Bert: Pre-training of deep bidirectional transformers for language
  understanding.
\newblock {\em arXiv preprint arXiv:1810.04805}, 2018.

\bibitem{kaplan2020scaling}
Jared Kaplan, Sam McCandlish, Tom Henighan, Tom~B Brown, Benjamin Chess, Rewon
  Child, Scott Gray, Alec Radford, Jeffrey Wu, and Dario Amodei.
\newblock Scaling laws for neural language models.
\newblock {\em arXiv preprint arXiv:2001.08361}, 2020.

\bibitem{bommasani2021opportunities}
Rishi Bommasani, Drew~A Hudson, Ehsan Adeli, Russ Altman, Simran Arora, Sydney
  von Arx, Michael~S Bernstein, Jeannette Bohg, Antoine Bosselut, Emma
  Brunskill, et~al.
\newblock On the opportunities and risks of foundation models.
\newblock {\em arXiv preprint arXiv:2108.07258}, 2021.

\bibitem{beltagy2019scibert}
Iz~Beltagy, Kyle Lo, and Arman Cohan.
\newblock Scibert: A pretrained language model for scientific text.
\newblock {\em arXiv preprint arXiv:1903.10676}, 2019.

\bibitem{cohan2020specter}
Arman Cohan, Sergey Feldman, Iz~Beltagy, Doug Downey, and Daniel~S Weld.
\newblock Specter: Document-level representation learning using
  citation-informed transformers.
\newblock {\em arXiv preprint arXiv:2004.07180}, 2020.

\bibitem{singh2022scirepeval}
Amanpreet Singh, Mike D'Arcy, Arman Cohan, Doug Downey, and Sergey Feldman.
\newblock Scirepeval: A multi-format benchmark for scientific document
  representations.
\newblock {\em arXiv preprint arXiv:2211.13308}, 2022.

\bibitem{nadkarni2021scientific}
Rahul Nadkarni, David Wadden, Iz~Beltagy, Noah~A Smith, Hannaneh Hajishirzi,
  and Tom Hope.
\newblock Scientific language models for biomedical knowledge base completion:
  an empirical study.
\newblock {\em arXiv preprint arXiv:2106.09700}, 2021.

\bibitem{gupta2022matscibert}
Tanishq Gupta, Mohd Zaki, and NM~Anoop Krishnan.
\newblock Matscibert: A materials domain language model for text mining and
  information extraction.
\newblock {\em npj Computational Materials}, 8(1):102, 2022.

\bibitem{taylor2022galactica}
Ross Taylor, Marcin Kardas, Guillem Cucurull, Thomas Scialom, Anthony
  Hartshorn, Elvis Saravia, Andrew Poulton, Viktor Kerkez, and Robert Stojnic.
\newblock Galactica: A large language model for science.
\newblock {\em arXiv preprint arXiv:2211.09085}, 2022.

\bibitem{azerbayev2023llemma}
Zhangir Azerbayev, Hailey Schoelkopf, Keiran Paster, Marco Dos~Santos, Stephen
  McAleer, Albert~Q. Jiang, Jia Deng, Stella Biderman, and Sean Welleck.
\newblock Llemma: An open language model for mathematics.
\newblock {\em arXiv preprint arXiv:2310.06786}, 2023.

\bibitem{xie2023darwin}
Tong Xie, Yuwei Wan, Wei Huang, Zhenyu Yin, Yixuan Liu, Shaozhou Wang, Qingyuan
  Linghu, Chunyu Kit, Clara Grazian, Wenjie Zhang, et~al.
\newblock Darwin series: Domain specific large language models for natural
  science.
\newblock {\em arXiv preprint arXiv:2308.13565}, 2023.

\bibitem{luo2022biogpt}
Renqian Luo, Liai Sun, Yingce Xia, Tao Qin, Sheng Zhang, Hoifung Poon, and
  Tie-Yan Liu.
\newblock Biogpt: generative pre-trained transformer for biomedical text
  generation and mining.
\newblock {\em Briefings in Bioinformatics}, 23(6):bbac409, 2022.

\bibitem{yang2022gatortron}
Xi~Yang, Aokun Chen, Nima PourNejatian, Hoo~Chang Shin, Kaleb~E Smith,
  Christopher Parisien, Colin Compas, Cheryl Martin, Mona~G Flores, Ying Zhang,
  et~al.
\newblock Gatortron: A large clinical language model to unlock patient
  information from unstructured electronic health records.
\newblock {\em arXiv preprint arXiv:2203.03540}, 2022.

\bibitem{deng2023learning}
Cheng Deng, Tianhang Zhang, Zhongmou He, Qiyuan Chen, Yuanyuan Shi, Le~Zhou,
  Luoyi Fu, Weinan Zhang, Xinbing Wang, Chenghu Zhou, et~al.
\newblock Learning a foundation language model for geoscience knowledge
  understanding and utilization.
\newblock {\em arXiv preprint arXiv:2306.05064}, 2023.

\bibitem{li2023llava}
Chunyuan Li, Cliff Wong, Sheng Zhang, Naoto Usuyama, Haotian Liu, Jianwei Yang,
  Tristan Naumann, Hoifung Poon, and Jianfeng Gao.
\newblock Llava-med: Training a large language-and-vision assistant for
  biomedicine in one day.
\newblock {\em arXiv preprint arXiv:2306.00890}, 2023.

\bibitem{tu2023towards}
Tao Tu, Shekoofeh Azizi, Danny Driess, Mike Schaekermann, Mohamed Amin,
  Pi-Chuan Chang, Andrew Carroll, Chuck Lau, Ryutaro Tanno, Ira Ktena, et~al.
\newblock Towards generalist biomedical ai.
\newblock {\em arXiv preprint arXiv:2307.14334}, 2023.

\bibitem{takeda2023foundation}
Seiji Takeda, Akihiro Kishimoto, Lisa Hamada, Daiju Nakano, and John~R Smith.
\newblock Foundation model for material science.
\newblock In {\em Proceedings of the AAAI Conference on Artificial
  Intelligence}, volume~37, pages 15376--15383, 2023.

\bibitem{nguyen2023climax}
Tung Nguyen, Johannes Brandstetter, Ashish Kapoor, Jayesh~K Gupta, and Aditya
  Grover.
\newblock Climax: A foundation model for weather and climate.
\newblock {\em arXiv preprint arXiv:2301.10343}, 2023.

\bibitem{brown2020language}
Tom Brown, Benjamin Mann, Nick Ryder, Melanie Subbiah, Jared~D Kaplan, Prafulla
  Dhariwal, Arvind Neelakantan, Pranav Shyam, Girish Sastry, Amanda Askell,
  et~al.
\newblock Language models are few-shot learners.
\newblock {\em Advances in neural information processing systems},
  33:1877--1901, 2020.

\bibitem{ouyang2022training}
Long Ouyang, Jeffrey Wu, Xu~Jiang, Diogo Almeida, Carroll Wainwright, Pamela
  Mishkin, Chong Zhang, Sandhini Agarwal, Katarina Slama, Alex Ray, et~al.
\newblock Training language models to follow instructions with human feedback.
\newblock {\em Advances in Neural Information Processing Systems},
  35:27730--27744, 2022.

\bibitem{GPT4}
{OpenAI}.
\newblock Gpt-4.
\newblock \url{https://openai.com/research/gpt-4}, 2023.
\newblock Version of the Generative Pre-trained Transformer.

\bibitem{ChatGPT}
{OpenAI}.
\newblock Chatgpt.
\newblock \url{https://openai.com/chatgpt}, 2023.
\newblock Software available from https://openai.com/chatgpt.

\bibitem{bran2023chemcrow}
Andres~M Bran, Sam Cox, Andrew~D White, and Philippe Schwaller.
\newblock Chemcrow: Augmenting large-language models with chemistry tools.
\newblock {\em arXiv preprint arXiv:2304.05376}, 2023.

\bibitem{white2022large}
Andrew~D White, Glen~M Hocky, Heta~A Gandhi, Mehrad Ansari, Sam Cox, Geemi~P
  Wellawatte, Subarna Sasmal, Ziyue Yang, Kangxin Liu, Yuvraj Singh, et~al.
\newblock Do large language models know chemistry?
\newblock {\em ChemRxiv}, 2022.

\bibitem{hatakeyama2023prompt}
Kan Hatakeyama-Sato, Naoki Yamane, Yasuhiko Igarashi, Yuta Nabae, and Teruaki
  Hayakawa.
\newblock Prompt engineering of gpt-4 for chemical research: what can/cannot be
  done?
\newblock {\em ChemRxiv}, 2023.

\bibitem{jablonka202314}
Kevin~Maik Jablonka, Qianxiang Ai, Alexander Al-Feghali, Shruti Badhwar,
  Joshua~D Bocarsly, Andres~M Bran, Stefan Bringuier, L~Catherine Brinson,
  Kamal Choudhary, Defne Circi, et~al.
\newblock 14 examples of how llms can transform materials science and
  chemistry: a reflection on a large language model hackathon.
\newblock {\em Digital Discovery}, 2(5):1233--1250, 2023.

\bibitem{guo2023can}
Taicheng Guo, Kehan Guo, Bozhao Nan, Zhenwen Liang, Zhichun Guo, Nitesh~V
  Chawla, Olaf Wiest, and Xiangliang Zhang.
\newblock What can large language models do in chemistry? a comprehensive
  benchmark on eight tasks.
\newblock In {\em Thirty-seventh Conference on Neural Information Processing
  Systems Datasets and Benchmarks Track}, 2023.

\bibitem{zheng2023large}
Yizhen Zheng, Huan~Yee Koh, Jiaxin Ju, Anh~TN Nguyen, Lauren~T May, Geoffrey~I
  Webb, and Shirui Pan.
\newblock Large language models for scientific synthesis, inference and
  explanation.
\newblock {\em arXiv preprint arXiv:2310.07984}, 2023.

\bibitem{qian2023can}
Chen Qian, Huayi Tang, Zhirui Yang, Hong Liang, and Yong Liu.
\newblock Can large language models empower molecular property prediction?
\newblock {\em arXiv preprint arXiv:2307.07443}, 2023.

\bibitem{wysocka2023large}
Magdalena Wysocka, Oskar Wysocki, Maxime Delmas, Vincent Mutel, and Andre
  Freitas.
\newblock Large language models, scientific knowledge and factuality: A
  systematic analysis in antibiotic discovery.
\newblock {\em arXiv preprint arXiv:2305.17819}, 2023.

\bibitem{nori2023capabilities}
Harsha Nori, Nicholas King, Scott~Mayer McKinney, Dean Carignan, and Eric
  Horvitz.
\newblock Capabilities of gpt-4 on medical challenge problems.
\newblock {\em arXiv preprint arXiv:2303.13375}, 2023.

\bibitem{wang2023large}
Yuqing Wang, Yun Zhao, and Linda Petzold.
\newblock Are large language models ready for healthcare? a comparative study
  on clinical language understanding.
\newblock {\em arXiv preprint arXiv:2304.05368}, 2023.

\bibitem{singhal2023large}
Karan Singhal, Shekoofeh Azizi, Tao Tu, S~Sara Mahdavi, Jason Wei, Hyung~Won
  Chung, Nathan Scales, Ajay Tanwani, Heather Cole-Lewis, Stephen Pfohl, et~al.
\newblock Large language models encode clinical knowledge.
\newblock {\em Nature}, 620(7972):172--180, 2023.

\bibitem{bordt2023chatgpt}
Sebastian Bordt and Ulrike von Luxburg.
\newblock Chatgpt participates in a computer science exam.
\newblock {\em arXiv preprint arXiv:2303.09461}, 2023.

\bibitem{wu2023empirical}
Yiran Wu, Feiran Jia, Shaokun Zhang, Qingyun Wu, Hangyu Li, Erkang Zhu, Yue
  Wang, Yin~Tat Lee, Richard Peng, and Chi Wang.
\newblock An empirical study on challenging math problem solving with gpt-4.
\newblock {\em arXiv preprint arXiv:2306.01337}, 2023.

\bibitem{pursnani2023performance}
Vinay Pursnani, Yusuf Sermet, Musa Kurt, and Ibrahim Demir.
\newblock Performance of chatgpt on the us fundamentals of engineering exam:
  Comprehensive assessment of proficiency and potential implications for
  professional environmental engineering practice.
\newblock {\em Computers and Education: Artificial Intelligence}, page 100183,
  2023.

\bibitem{zheng2023can}
Mingkai Zheng, Xiu Su, Shan You, Fei Wang, Chen Qian, Chang Xu, and Samuel
  Albanie.
\newblock Can gpt-4 perform neural architecture search?
\newblock {\em arXiv preprint arXiv:2304.10970}, 2023.

\bibitem{zhang2023automl}
Shujian Zhang, Chengyue Gong, Lemeng Wu, Xingchao Liu, and Mingyuan Zhou.
\newblock Automl-gpt: Automatic machine learning with gpt.
\newblock {\em arXiv preprint arXiv:2305.02499}, 2023.

\bibitem{vijay2023prompt}
Vijay Viswanathan, Chenyang Zhao, Amanda Bertsch, Tongshuang Wu, and Graham
  Neubig.
\newblock Prompt2model: Generating deployable models from natural language
  instructions.
\newblock {\em arXiv preprint arXiv:2308.12261}, 2023.

\bibitem{wang2023survey}
Lei Wang, Chen Ma, Xueyang Feng, Zeyu Zhang, Hao Yang, Jingsen Zhang, Zhiyuan
  Chen, Jiakai Tang, Xu~Chen, Yankai Lin, et~al.
\newblock A survey on large language model based autonomous agents.
\newblock {\em arXiv preprint arXiv:2308.11432}, 2023.

\bibitem{bail2023can}
Christopher~A Bail.
\newblock Can generative ai improve social science?
\newblock {\em SocArXiv}, 2023.

\bibitem{ziems2023can}
Caleb Ziems, William Held, Omar Shaikh, Jiaao Chen, Zhehao Zhang, and Diyi
  Yang.
\newblock Can large language models transform computational social science?
\newblock {\em arXiv preprint arXiv:2305.03514}, 2023.

\bibitem{park2023generative}
Joon~Sung Park, Joseph~C O'Brien, Carrie~J Cai, Meredith~Ringel Morris, Percy
  Liang, and Michael~S Bernstein.
\newblock Generative agents: Interactive simulacra of human behavior.
\newblock {\em arXiv preprint arXiv:2304.03442}, 2023.

\bibitem{horton2023large}
John~J Horton.
\newblock Large language models as simulated economic agents: What can we learn
  from homo silicus?
\newblock Technical report, National Bureau of Economic Research, 2023.

\bibitem{korinek2023generative}
Anton Korinek.
\newblock Generative ai for economic research: Use cases and implications for
  economists.
\newblock {\em Journal of Economic Literature}, 61(4), 2023.

\bibitem{aher2023using}
Gati~V Aher, Rosa~I Arriaga, and Adam~Tauman Kalai.
\newblock Using large language models to simulate multiple humans and replicate
  human subject studies.
\newblock In {\em International Conference on Machine Learning}, pages
  337--371. PMLR, 2023.

\bibitem{alzaabi2023chatgpt}
Adhari AlZaabi, Amira ALamri, Halima Albalushi, Ruqaya Aljabri, and Abdulrahman
  AalAbdulsallam.
\newblock Chatgpt applications in academic research: A review of benefits,
  concerns, and recommendations.
\newblock {\em bioRxiv}, pages 2023--08, 2023.

\bibitem{elicit}
{Elicit}.
\newblock \url{https://elicit.org/}.
\newblock Accessed on 2023-04-06.

\bibitem{scispace}
{SCISPACE}.
\newblock \url{https://scispace.com/}.
\newblock Accessed on 2023-04-06.

\bibitem{transformer2022can}
Gpt Generative~Pretrained Transformer, Almira~Osmanovic Thunstr{\"o}m, and
  Steinn Steingrimsson.
\newblock Can gpt-3 write an academic paper on itself, with minimal human
  input?
\newblock {\em HAL Open Science}, 2022.

\bibitem{gao2023comparing}
Catherine~A Gao, Frederick~M Howard, Nikolay~S Markov, Emma~C Dyer, Siddhi
  Ramesh, Yuan Luo, and Alexander~T Pearson.
\newblock Comparing scientific abstracts generated by chatgpt to real abstracts
  with detectors and blinded human reviewers.
\newblock {\em NPJ Digital Medicine}, 6(1):75, 2023.

\bibitem{aydin2022openai}
{\"O}mer Ayd{\i}n and Enis Karaarslan.
\newblock Openai chatgpt generated literature review: Digital twin in
  healthcare.
\newblock {\em Available at SSRN 4308687}, 2022.

\bibitem{wexin2023can}
Weixin Liang, Yuhui Zhang, Hancheng Cao, Binglu Wang, Daisy Ding, Xinyu Yang,
  Kailas Vodrahalli, Siyu He, Daniel Smith, Yian Yin, Daniel McFarland, and
  James Zou.
\newblock {Can large language models provide useful feedback on research
  papers? A large-scale empirical analysis}.
\newblock In {\em arXiv preprint arXiv:2310.01783}, 2023.

\bibitem{liu2023reviewergpt}
Ryan Liu and Nihar~B Shah.
\newblock Reviewergpt? an exploratory study on using large language models for
  paper reviewing.
\newblock {\em arXiv preprint arXiv:2306.00622}, 2023.

\bibitem{robertson2023gpt4}
Zachary Robertson.
\newblock Gpt4 is slightly helpful for peer-review assistance: A pilot study.
\newblock {\em arXiv preprint arXiv:2307.05492}, 2023.

\bibitem{hosseini2023fighting}
Mohammad Hosseini and Serge~PJM Horbach.
\newblock Fighting reviewer fatigue or amplifying bias? considerations and
  recommendations for use of chatgpt and other large language models in
  scholarly peer review.
\newblock {\em Research Integrity and Peer Review}, 8(1):4, 2023.

\bibitem{krenn2022scientific}
Mario Krenn, Robert Pollice, Si~Yue Guo, Matteo Aldeghi, Alba Cervera-Lierta,
  Pascal Friederich, Gabriel dos Passos~Gomes, Florian H{\"a}se, Adrian Jinich,
  AkshatKumar Nigam, et~al.
\newblock On scientific understanding with artificial intelligence.
\newblock {\em Nature Reviews Physics}, 4(12):761--769, 2022.

\bibitem{arrieta2020explainable}
Alejandro~Barredo Arrieta, Natalia D{\'\i}az-Rodr{\'\i}guez, Javier Del~Ser,
  Adrien Bennetot, Siham Tabik, Alberto Barbado, Salvador Garc{\'\i}a, Sergio
  Gil-L{\'o}pez, Daniel Molina, Richard Benjamins, et~al.
\newblock Explainable artificial intelligence (xai): Concepts, taxonomies,
  opportunities and challenges toward responsible ai.
\newblock {\em Information fusion}, 58:82--115, 2020.

\bibitem{openinterpreter}
KillianLucas.
\newblock Open interpreter, 2023.
\newblock Accessed: 2023-09-24, License: MIT.

\bibitem{openai_chatgpt_plugins_code_interpreter_2023}
{OpenAI}.
\newblock Chatgpt plugins - code interpreter.
\newblock \url{https://openai.com/blog/chatgpt-plugins#code-interpreter}, 2023.
\newblock Accessed: 2023-12-03.

\bibitem{nakano2021webgpt}
Reiichiro Nakano, Jacob Hilton, Suchir Balaji, Jeff Wu, Long Ouyang, Christina
  Kim, Christopher Hesse, Shantanu Jain, Vineet Kosaraju, William Saunders,
  et~al.
\newblock Webgpt: Browser-assisted question-answering with human feedback.
\newblock {\em arXiv preprint arXiv:2112.09332}, 2021.

\bibitem{act1}
Adept.
\newblock Act-1: Transformer for actions, 2022.
\newblock Accessed: 2023-09-25.

\bibitem{xi2023rise}
Zhiheng Xi, Wenxiang Chen, Xin Guo, Wei He, Yiwen Ding, Boyang Hong, Ming
  Zhang, Junzhe Wang, Senjie Jin, Enyu Zhou, et~al.
\newblock The rise and potential of large language model based agents: A
  survey.
\newblock {\em arXiv preprint arXiv:2309.07864}, 2023.

\bibitem{hutter2019automated}
Frank Hutter, Lars Kotthoff, and Joaquin Vanschoren.
\newblock {\em Automated machine learning: methods, systems, challenges}.
\newblock Springer Nature, 2019.

\bibitem{bischl2023hyperparameter}
Bernd Bischl, Martin Binder, Michel Lang, Tobias Pielok, Jakob Richter, Stefan
  Coors, Janek Thomas, Theresa Ullmann, Marc Becker, Anne-Laure Boulesteix,
  et~al.
\newblock Hyperparameter optimization: Foundations, algorithms, best practices,
  and open challenges.
\newblock {\em Wiley Interdisciplinary Reviews: Data Mining and Knowledge
  Discovery}, 13(2):e1484, 2023.

\bibitem{lindauer2020best}
Marius Lindauer and Frank Hutter.
\newblock Best practices for scientific research on neural architecture search.
\newblock {\em The Journal of Machine Learning Research}, 21(1):9820--9837,
  2020.

\bibitem{white2023neural}
Colin White, Mahmoud Safari, Rhea Sukthanker, Binxin Ru, Thomas Elsken, Arber
  Zela, Debadeepta Dey, and Frank Hutter.
\newblock Neural architecture search: Insights from 1000 papers.
\newblock {\em arXiv preprint arXiv:2301.08727}, 2023.

\bibitem{kreuzberger2023machine}
Dominik Kreuzberger, Niklas K{\"u}hl, and Sebastian Hirschl.
\newblock Machine learning operations (mlops): Overview, definition, and
  architecture.
\newblock {\em IEEE Access}, 2023.

\bibitem{yuan2022can}
Weizhe Yuan, Pengfei Liu, and Graham Neubig.
\newblock Can we automate scientific reviewing?
\newblock {\em Journal of Artificial Intelligence Research}, 75:171--212, 2022.

\bibitem{yuan2022kid}
Weizhe Yuan and Pengfei Liu.
\newblock Kid-review: Knowledge-guided scientific review generation with oracle
  pre-training.
\newblock In {\em Proceedings of the AAAI Conference on Artificial
  Intelligence}, volume~36, pages 11639--11647, 2022.

\bibitem{wang2020reviewrobot}
Qingyun Wang, Qi~Zeng, Lifu Huang, Kevin Knight, Heng Ji, and Nazneen~Fatema
  Rajani.
\newblock Reviewrobot: Explainable paper review generation based on knowledge
  synthesis.
\newblock {\em arXiv preprint arXiv:2010.06119}, 2020.

\bibitem{schulz2022future}
Robert Schulz, Adrian Barnett, Ren{\'e} Bernard, Nicholas~JL Brown, Jennifer~A
  Byrne, Peter Eckmann, Ma{\l}gorzata~A Gazda, Halil Kilicoglu, Eric~M Prager,
  Maia Salholz-Hillel, et~al.
\newblock Is the future of peer review automated?
\newblock {\em BMC Research Notes}, 15(1):1--5, 2022.

\bibitem{zhao2022reviewer}
Xiquan Zhao and Yangsen Zhang.
\newblock Reviewer assignment algorithms for peer review automation: A survey.
\newblock {\em Information Processing \& Management}, 59(5):103028, 2022.

\end{thebibliography}

\end{document}